\documentclass{article}

\usepackage{microtype}
\usepackage{graphicx}
\usepackage{subfigure}
\usepackage{booktabs} 

\usepackage[colorlinks=true, linkcolor=black, citecolor=black, urlcolor=black]{hyperref}

\usepackage[ruled,vlined,linesnumbered,algo2e]{algorithm2e}

\usepackage[accepted]{icml2025}

\usepackage{amsmath}
\usepackage{amssymb}
\usepackage{mathtools}
\usepackage{amsthm}
\usepackage{color}
\usepackage{booktabs}

\usepackage{blindtext}
\usepackage{enumerate}
\usepackage{graphicx}
\usepackage{amsfonts, amsmath, bm, amssymb}
\usepackage{dsfont}
\usepackage{wrapfig}
\usepackage{subcaption}
\usepackage{enumitem}
\usepackage{booktabs}
\usepackage{multirow}
\usepackage{xfrac}
\usepackage{makecell}
\usepackage{overpic}
\newcommand\numberthis{\addtocounter{equation}{1}\tag{\theequation}}
\usepackage{mwe}
\usepackage{listings}
\definecolor{darkgreen}{RGB}{0, 128, 27}
\usepackage{placeins}
\usepackage{thm-restate}

\def\conv{\hspace{-0.1em}*\hspace{-0.1em}}

\newcommand{\zero}{\mathbf{0}}

\newcommand{\diff}{\,\mathrm{d}}

\newcommand{\inner}[2]{\left\langle #1, #2 \right\rangle}

\newcommand{\KL}[2]{D_{\mathrm{KL}}({#1} \; \| \; {#2})}

\usepackage{xspace}
\makeatletter
\DeclareRobustCommand\onedot{\futurelet\@let@token\@onedot}
\def\@onedot{\ifx\@let@token.\else.\null\fi\xspace}

\makeatother

\newcommand{\Bc}{\mathcal{B}}

\newcommand{\Dc}{\mathcal{D}}
\newcommand{\Ec}{\mathcal{E}}

\newcommand{\Lc}{\mathcal{L}}

\newcommand{\Nc}{\mathcal{N}}
\newcommand{\Oc}{\mathcal{O}}

\newcommand{\Tc}{\mathcal{T}}

\newcommand{\Yc}{\mathcal{Y}}

\newcommand{\Eb}{\mathbb{E}}

\newcommand{\Rb}{\mathbb{R}}

\newcommand{\bv}{\mathbf{b}}

\newcommand{\fv}{\mathbf{f}}

\newcommand{\sv}{\mathbf{s}}

\newcommand{\uv}{\mathbf{u}}
\newcommand{\wv}{\mathbf{w}}
\newcommand{\xv}{\mathbf{x}}
\newcommand{\yv}{\mathbf{y}}

\newcommand{\Iv}{\mathbf{I}}

\newcommand{\epsilonv   }{\boldsymbol \epsilon   }

\newcommand{{\phiv}     }{\boldsymbol \phi       }

\ifx\BlackBox\undefined
\newcommand{\BlackBox}{\rule{1.5ex}{1.5ex}}  
\fi
\ifx\QED\undefined
\def\QED{~\rule[-1pt]{5pt}{5pt}\par\medskip}
\fi
\ifx\proof\undefined
\newenvironment{proof}{\par\noindent{\em Proof:\ }}{\hfill\BlackBox\\}
\fi
\ifx\theorem\undefined
\newtheorem{theorem}{Theorem}
\fi
\ifx\example\undefined
\newtheorem{example}{Example}
\fi
\ifx\property\undefined
\newtheorem{property}{Property}
\fi
\ifx\lemma\undefined
\newtheorem{lemma}{Lemma}
\fi
\ifx\proposition\undefined
\newtheorem{proposition}{Proposition}
\fi
\ifx\fact\undefined
\newtheorem{fact}{Fact}
\fi
\ifx\remark\undefined
\newtheorem{remark}{Remark}
\fi
\ifx\corollary\undefined
\newtheorem{corollary}{Corollary}
\fi
\ifx\definition\undefined
\newtheorem{definition}{Definition}
\fi
\ifx\conjecture\undefined
\newtheorem{conjecture}{Conjecture}
\fi
\ifx\axiom\undefined
\newtheorem{axiom}[theorem]{Axiom}
\fi
\ifx\claim\undefined
\newtheorem{claim}[theorem]{Claim}
\fi
\ifx\assumption\undefined
\newtheorem{assumption}{Assumption}
\fi
\ifx\question\undefined
\newtheorem{question}{Question}
\fi
\ifx\problem\undefined
\newtheorem{problem}{Problem}
\fi

\usepackage[capitalize,noabbrev]{cleveref}
\usepackage{textcomp} 
\crefformat{prop}{Prop~#2#1#3}
\crefformat{figure}{Fig~#2#1#3}
\crefformat{lem}{Lem~#2#1#3}
\crefformat{lemma}{Lem~#2#1#3}
\crefformat{section}{Sec~#2#1#3}
\crefformat{equation}{Eq~(#2#1#3)}
\crefformat{algorithm}{Alg~#2#1#3}
\crefformat{fact}{Fact~#1}
\crefformat{chapter}{Chapter~#2#1#3}
\crefalias{lem}{Lemma}
\crefformat{theorem}{Thm~#1}
\crefformat{corollary}{Cor~#2#1#3}
\crefformat{proposition}{Prop~#2#1#3}
\Crefname{appendix}{Appx}{Appx}

\usepackage[textsize=tiny]{todonotes}

\icmltitlerunning{SFBD: 
Training Diffusion Models with Finite Noisy Datasets}

\begin{document}

\twocolumn[
\icmltitle{Stochastic Forward–Backward Deconvolution: \\
Training Diffusion Models with Finite Noisy Datasets}



\icmlsetsymbol{equal}{*}

\begin{icmlauthorlist}
\icmlauthor{Haoye Lu}{sch,vi}
\icmlauthor{Qifan Wu}{sch}
\icmlauthor{Yaoliang Yu}{sch,vi}
\end{icmlauthorlist}

\icmlaffiliation{vi}{Vector Institute, Canada}
\icmlaffiliation{sch}{Cheriton School of Computer Science, University of Waterloo, Canada}

\icmlcorrespondingauthor{Haoye Lu}{haoye.lu@uwaterloo.ca}

\icmlkeywords{Machine Learning, ICML}

\vskip 0.3in
]



\printAffiliationsAndNotice{}  

\begin{abstract}
Recent diffusion-based generative models achieve remarkable results by training on massive datasets, yet this practice raises concerns about memorization and copyright infringement. A proposed remedy is to train exclusively on noisy data with potential copyright issues, ensuring the model never observes original content. However, through the lens of deconvolution theory, we show that although it is theoretically feasible to learn the data distribution from noisy samples, the practical challenge of collecting sufficient samples makes successful learning nearly unattainable. To overcome this limitation, we propose to pretrain the model with a small fraction of clean data to guide the deconvolution process. Combined with our Stochastic Forward--Backward Deconvolution (SFBD) method, we attain FID  $6.31$ on CIFAR-10 with just $4\%$ clean images (and $3.58$ with $10\%$). We also provide theoretical guarantees that SFBD learns the true data distribution. These results underscore the value of limited clean pretraining, or pretraining on similar datasets. Empirical studies further validate and enrich our findings. 
\end{abstract}


\section{Introduction}
\label{sec:intro}
Diffusion-based generative models \citep{DicksteinWMG2015,HoJA2020, SongME2021, SongDCKKEP2021, SongDCS2023} have gained increasing attention. Nowadays, it is considered one of the most powerful frameworks for learning high-dimensional distributions and we have witnessed many impressive breakthroughs  \citep{CroitoruHIS2023} in generating images \citep{HoJA2020, SongME2021, SongDCKKEP2021, RombachBLEO2022, SongDCS2023}, audios \citep{kong2021diffwave, YangYWWWZY2023} and videos \citep{HoSGCNF2022}.

Due to some inherent properties, diffusion models are relatively easier to train. This unlocks the possibility of training very large models on web-scale data, which has been shown to be critical to train powerful models. This paradigm has recently led to impressive advances in image generation, as demonstrated by cutting-edge models like Stable Diffusion (-XL) \citep{RombachBLEO2022, PodellELBDMPR2023} and DALL-E (2, 3) \citep{BGJBWLOZLG2023}. However, despite their success, the reliance on extensive web-scale data introduces challenges. The complexities of the datasets at such a scale often result in the inclusion of copyrighted content. Furthermore, diffusion models exhibit a greater tendency than earlier generative approaches, such as Generative Adversarial Networks (GANs) \citep{GoodfellowPMXWOCB2014, GoodfellowPMXWOCB2020}, to memorize training examples. This can lead to the replication of parts or even entire images from their training sets \citep{CarliniHNJSTBIDW2023, SomepalliSGGG2023}.

A recently proposed approach to address memorization and copyright concerns involves training (or fine-tuning) diffusion models using corrupted samples~\cite{DarasSDGDK2023, SomepalliSGGG2023, DarasA2023, DarasDD2024}. In this framework, the model is never exposed to the original samples during training. Instead, these samples undergo a known non-invertible corruption process, such as adding independent Gaussian noise to each pixel in image datasets. This ensures that the model cannot memorize or reproduce the original content, as the corruption process is irreversible for individual samples. 
 
Interestingly, under mild assumptions, certain non-invertible corruption processes, such as Gaussian noise injection, create a mathematical bijection between the noisy and original distributions. Thus, in theory, a generative model can learn the original distribution using only noisy samples \citep{BoraPD2018}. Building on this concept, \citet{DarasDD2024} demonstrated that when an image is corrupted via a forward diffusion up to a specific noise level $\sigma$, diffusion models can recover distributions at noise levels below $\sigma$ by enforcing consistency constraints \citep{DarasDDD2023}.

While \citet{DarasDD2024} empirically showed that their approach could be used to fine-tune Stable Diffusion XL~\citep{PodellELBDMPR2023} using noisy images with a heuristic consistency loss, they did not explore whether a diffusion model can be successfully trained solely with noisy images. Moreover, the effectiveness of the consistency loss in such scenarios remains an open question. 

In this paper, we address these questions by connecting the task of estimating the original distribution from noisy samples to the well-studied density deconvolution problem \citep{Meister2009}. Through the lens of deconvolution theory, we establish that the optimal convergence rate for estimating the data density is $\Oc(\log n)^{-2}$ when $n$ noisy samples are generated via a forward diffusion process. This pessimistic rate suggests that while it is theoretically feasible to learn the data distribution from noisy samples, the practical challenge of collecting sufficient samples makes successful learning nearly unattainable. Our empirical studies further validate this theoretical insight and suggest the inefficiency of the current consistency loss outside the regime of fine-tuning latent diffusion models.

To address the poor convergence rate in training diffusion models with noisy data, we propose pretraining models on a small subset of copyright-free clean data as an effective solution. Since the current consistency loss remains ineffective even with pretraining, we propose a new deconvolution method, Stochastic Forward–Backward Deconvolution (SFBD, pronounced \texttt{sofabed}), that is fully compatible with the existing diffusion training framework. Experimentally, we achieve an FID of 6.31 with just 4\% clean images on CIFAR-10 and 3.58 with 10\% clean images. Our theoretical results ensure that the learnt distribution converges to the true data distribution and justify the necessity of pretraining. Furthermore, our results suggest that models can be pretrained using datasets with similar features when clean, copyright-free data are unavailable. Ablation studies provide additional evidence supporting our claims. Code for the empirical study is available at: \href{https://github.com/watml/SFBD.git}{github.com/watml/SFBD}.

\section{Related Work}
\label{sec:related}
The rise of large diffusion models trained on massive datasets has sparked growing concerns about copyright infringement and memorization of training data \citep{CarliniHNJSTBIDW2023, SomepalliSGGG2023}. While differential privacy (DP) has been explored as a mitigation strategy \citep{AbadiCGMMTZ2016, XieLWWR2018, DockhornCVK2023}, it often presents practical challenges. Notably, DP can require users to share their original data with a central server for training unless local devices have sufficient computational power for backpropagation.


In contrast, training on corrupted data provides a compelling alternative, allowing users to contribute without exposing their original data. By sharing only non-invertible, corrupted versions, sensitive information remains on users’ devices, eliminating the need to transmit original data.

Learning generative models from corrupted data poses a significant challenge, as the model must reconstruct the underlying data distribution from incomplete or noisy information. In their work on AmbientGAN, \citet{BoraPD2018} showed that it is empirically feasible to train GANs using corrupted images. They also provided a theoretical guarantee that, with a sufficient number of corrupted samples generated by randomly blacking out pixels, the learned distribution converges to the true data distribution. Building on this, \citet{WangZHCZ2023} demonstrated a closely related result: under certain weak assumptions, if the model-generated fake samples and the corrupted true samples share the same distribution after undergoing identical corruption, then the fake data distribution aligns perfectly with the true data distribution. Their analysis applies to scenarios where corruption is implemented via a forward diffusion process but does not address cases where the two corrupted distributions are similar but not identical -- a case we explore in \cref{prop:conv_identify} below.


Inspired by the success of training GANs using corrupted data, \citet{DarasSDGDK2023, AaliAKT2023, DarasA2023, BaiWCS2024, DarasDD2024} demonstrated the feasibility of training diffusion models with corrupted data. Notably, \citet{DarasDD2024} showed that when corruption is performed by a forward diffusion process, the marginal distribution at one time step determines the distributions at other time steps, all of which must satisfy certain consistency constraints. Building on this, they showed that if a model learns distributions above the corruption noise level, it can infer those below the noise level by adhering to these constraints. To enforce this, they introduced a consistency loss to improve compliance with the constraints, though its effectiveness was demonstrated only in fine-tuning latent diffusion models.

Outside the field of machine learning, the problem of estimating the original distribution from noisy samples has traditionally been addressed through density deconvolution~\citep{Meister2009}. This research area aims to recover the distribution of error-free data from noise-contaminated observations. Most existing deconvolution methods are limited to the univariate setting \citep{CarrollHall88, Zhang1990, Fan91, CordyT1997, DelaigleH2008, MeisterN2010, LouniciN2011, Guan2021}, with only a few approaches extending to the multivariate case. These multivariate techniques typically rely on normal mixture models \citep{BovyHR2011, SarkarPCMC2018} or kernel smoothing methods \citep{Masry1993, LepskiW2019}. Integrating these theoretical insights into modern generative model frameworks remains a significant challenge. However, by reinterpreting generative models trained on noisy data through the lens of deconvolution theory, we can gain a deeper understanding of their fundamental limitations and capabilities, as they inherently address the deconvolution problem.

A very recent study by \citet{DarasCD25}, using Gaussian Mixture Models, also highlights the challenge of training diffusion models with only noisy samples and shows that adding a few clean samples can significantly improve performance. The convergence of conclusions from fundamentally different approaches reinforces the findings of both works. Methodologically, \citet{DarasCD25} apply Tweedie’s formula (consistency constraint) to recover the clean distribution. In contrast, ours introduces a novel forward-backward deconvolution strategy, offering a fresh perspective without the heavy computational cost of enforcing consistency.

Beyond diffusion model training, \citet{BieKS2022} and \citet{DavidBCKS2023} demonstrate that even a small amount of clean public data can substantially reduce the sample complexity in differentially private (DP) estimation when learning from sensitive data. Although motivated by a different goal, \citet{NieGHXVA2022} propose DiffPure, an algorithm that leverages a pretrained diffusion model to remove adversarial perturbations via a forward–backward diffusion process. While DiffPure and SFBD share a similar structure, they serve distinct purposes: DiffPure assumes a well-trained model to purify data, whereas SFBD is designed to train the diffusion model itself.



\section{Prelimilaries}
\label{sec:prel}
In this section we recall diffusion models, the density deconvolution problem and the consistency constraints.
\subsection{Diffusion Models}
Diffusion models generate data by progressively adding Gaussian noise to input data and then reversing this process through sequential denoising steps to sample from noise. Given distribution $p_0$ on $\Rb^d$, the forward perturbation is specified by a stochastic differential equation (SDE):
\begin{align}
	\diff \xv_t = g(t) \diff \wv_t, ~~\textrm{~ $t\in [0, T]$,} \label{eq:fwd_diff}
\end{align}
$\xv_0 \sim p_0$, $T$ is a fixed positive constant and $g(t)$ is a scalar function.  $\{\wv_t\}_{t\in [0,T]}$ is the standard Brownian motion. 

\cref{eq:fwd_diff} induces a transition kernel $p_{t|s}(\xv_t | \xv_s)$ for $0 \leq s \leq t \leq T$, which is Gaussian and its mean and covariance matrix can be computed in closed form \citep[Eqs 4.23 and 5.51]{SarkkaSolin2019}. In particular, for $s = 0$, we write 
\begin{align}
    p_{t|0}(\xv_t | \xv_0) = \Nc(\xv_0, \sigma_t^2 \Iv),
\end{align}  
for all $ t \in [0, T] $, where we set $g(t) = (\frac{\diff \sigma^2_t}{\diff t})^{\sfrac{1}{2}}$. 
When $\sigma_T^2$ is very large, $\xv_T$ can be approximately regarded as a sample from $\Nc(\zero, \sigma_T^2 \Iv)$. Let $ p_t(\xv_t) = \int p_{t|0}(\xv_t \vert \xv_0) \, p_0(\xv_0) \, \mathrm{d} \xv_0 $ denote the marginal distribution of $\xv_t$, where we have $ p_T \approx \mathcal{N}(\mathbf{0}, \sigma_T^2 \mathbf{I}) $. \citet{anderson1982} showed that backward SDE
\begin{align}
	\diff \xv_t = - g(t)^2 \, \nabla \log p_t(\xv_t) \diff t + g(t) \diff \bar\wv_t, ~\xv_T \sim p_T \label{eq:anderson_bwd}
\end{align}
has a transition kernel that matches the posterior distribution of the forward process, $p_{s|t}(\xv_s | \xv_t) = \tfrac{p_{t|s}(\xv_t | \xv_s)  p_s(\xv_s)}{p_t(\xv_t)}$ for $s \leq t$ in $[0,T]$. Thus, the backward SDE preserves the same marginal distributions as the forward process. Here, $\bar{\wv}_t$ represents a standard Wiener process with time flowing backward from $T$ to $0$, while $\nabla \log p_t(\xv_t)$ denotes the score function of the distribution $p_t(\xv_t)$.  With a well-trained network $\sv_{\phiv}(\xv_t, t) \approx \nabla \log p_t(\xv_t)$, we substitute it into \cref{eq:anderson_bwd} and solve the SDE backward from $\tilde\xv_T \sim \Nc(\zero, \sigma_T^2 \Iv)$. The resulting $\tilde\xv_0$ then serves as an approximate sample of $p_0$.

To train $\sv_{\phiv}$ to estimate the score, let $\Tc$ be a predefined sampler of $t \in [0, T]$ and $w(t)$ be a weight function. The network $\sv_{\phiv}$ can be effectively trained via the conditional score-matching loss \citep{SongDCKKEP2021}:
\begin{align*}
	\Lc_s(\phiv) \hspace{-0.3em} = \hspace{-0.4em} \underset{t \sim \Tc}{\Eb}\, \underset{p_0}{\Eb} \,  \underset{p_{t|0}}{\Eb} \left[w(t) \|\sv_{\phiv}(\xv_t, t) - \nabla \log p_{t\vert 0}(\xv_t \vert \xv_0)\|^2 \right]  
\end{align*}
Instead, we may first train a denoiser $D_{\phiv}(\xv, t)$ to estimate $\Eb[\xv_0 \vert \xv_t]$ by minimizing \citep{KarrasAAL22}
\begin{align}
	\Lc_d(\phiv) \hspace{-0.3em} = \hspace{-0.4em} \underset{t \sim \Tc}{\Eb}\, \underset{p_0}{\Eb} \,  \underset{p_{t|0}}{\Eb} \left[w(t) \|D_{\phiv}(\xv_t, t) - \xv_0 \|^2 \right] \label{eq:denoiser_loss}
\end{align}
then estimate 
\begin{align}
	\nabla \log p_t(\xv_t) = \frac{\Eb[\xv_0 \vert \xv_t] - \xv_t}{\sigma_t^2} \approx \frac{D_{\phiv}(\xv_t, t) - \xv_t}{\sigma_t^2}. \label{eq:score_repara_in_denoiser}
\end{align}

%
%
%
%
%
%
%
%
%

\subsection{Density Deconvolution Problems}
\label{sec:prel:deconv_prob}

Classical deconvolution problems arise in scenarios where data are corrupted due to significant measurement errors, and the goal is to estimate the underlying data distribution. Specifically, let the corrupted samples $\mathcal{Y} = \{\yv^{(i)}\}_{i=1}^{n}$ be generated by the process:
\begin{align}
	\yv^{(i)} = \xv^{(i)} + \epsilonv^{(i)}, \label{eq:gen_conv_samples}
\end{align}
where $\xv^{(i)}$ and $\epsilonv^{(i)}$ are independent random variables. Here, $\xv^{(i)}$ is drawn from an unknown distribution with density $p_{\rm data}$, and $\epsilonv^{(i)}$ is sampled from a \emph{known} error distribution with density $h$. It can be shown that the corrupted samples $\yv^{(i)}$ follow a distribution with density $p_{\rm data} \conv h$, where $\conv$ denotes the convolution operator. We provide more details in \cref{appx:intro:density_deconv}.

The objective of the (density) deconvolution problem is to estimate the density of $p_{\rm data}$ using the observed data $\mathcal{Y}$, which is sampled from the convoluted distribution $p_{\rm data} \conv h$. In essence, deconvolution reverses the density convolution process, hence the name of the problem.

To assess the quality of an estimator $\hat{p}(\cdot; \mathcal{Y})$ of $p_{\rm data}$ based on $\mathcal{Y}$, the mean integrated squared error (MISE) is commonly used. MISE is defined as:
\begin{align}
	\textrm{MISE}( \hat p, p_\textrm{data}) = \Eb_{\Yc} \int_{\Rb^d} \big|\,                                                            \hat p(\xv ; \Yc) - p_\textrm{data}(\xv) \, \big|^2 \diff \xv. \label{eq:def_MISE}
\end{align}
In this paper, we focus on a corruption process implemented via forward diffusion as described in \cref{eq:fwd_diff}. Consequently, unless otherwise stated, in the rest of this work, we assume the error distribution $h$ is Gaussian $\mathcal{N}(\mathbf{0}, \sigma_\zeta^2\mathbf{I})$ with a given and fixed $\zeta \in (0, T)$.

To see why we could identify an original distribution $p$ through $p \conv h$, let $\Phi_p(\uv) = \Eb_p[\exp(i \, \uv^\top \xv)]$ for $\uv \in \Rb^d$ be the characteristic function of $p$. Then,

\begin{restatable}{proposition}{PROPCONVINDENTIFY}
\label{prop:conv_identify}
	Let $p$ and $q$ be two distributions defined on $\Rb^d$. For all $\uv \in \Rb^d$, 
	\begin{align*}
			|\Phi_p(\uv)\hspace{-0.1em} -\hspace{-0.1em} \Phi_q(\uv)|\hspace{-0.15em}\leq\hspace{-0.15em} \exp\hspace{-0.1em}\big(\frac{\sigma_\zeta^2}{2} \|\uv\|^2 \big) \sqrt{2 \,D_{\mathrm{KL}}(p \conv h\|q\conv h)}.
	\end{align*}
\end{restatable}
(All proofs are deferred to the appendix.) This result shows if two distributions $p$ and $q$ are similar after being convoluted with $h$, they must have similar characteristic functions and thus similar distribution. In particular, when $p \conv h = q \conv h$, then $p = q$, the case also discussed in \citet[Thm 2]{WangZHCZ2023}. As a result, whenever we could find $q$ satisfying $p_\text{data} \conv h = q \conv h$, we can conclude $p_\text{data} = q$.   

\subsection{Deconvolution through the Consistency Constraints}
\label{prel:deconv_through_consist_const}
While \cref{prop:conv_identify} shows it is possible to train a generative model using noisy samples, it remains a difficult question of how to use noisy samples to train a diffusion model to generate clean samples \emph{effectively}.

The question was partially addressed by \citet{DarasDD2024} through the consistency property \citep{DarasDDD2023}. In particular, since we have access to the noisy samples $\xv_\zeta$ from $p_\text{data} \conv h$, we can use them to train a network $\sv_{\phiv}(\xv_t, t)$ to approximate $\nabla \log p_t(\xv_t)$ for $t > \zeta$ through a modified score matching loss, which is referred as ambient score matching (ASM), denoted by $\Lc_\text{ASM}({\phiv})$. In their implementation, $\sv_{\phiv}(\xv_t, t)$ is parameterized by $\frac{D_{\phiv}(\xv_t, t) - \xv_t}{\sigma_t^2}$, where $D_{\phiv}(\xv_t, t)$ is trained to approximate $\Eb[\xv_0 \vert \xv_t]$. 

In contrast, for $t \leq \zeta$, score-matching is no longer applicable. Instead, \citet{DarasDD2024} propose that $D_{\phiv}(\xv_t, t)$ should obey the consistency property: 
\begin{align}
	\Eb[\xv_0|\xv_s] =  {\Eb}_{p_{r\vert s}} \big[\Eb[\xv_0|\xv_r] \big], \text{ for $0 \leq r \leq s \leq T$}
\end{align} 
by jointly minimizing the \textit{consistency loss}:
\begin{align}
	\Lc_\text{con}({\phiv}, r, s) \hspace{-0.12em}=\hspace{-0.12em} \Eb_{p_s} \big\|D_{\phiv}(\xv_s, s) \hspace{-0.12em} - \hspace{-0.12em} \Eb_{p_{r|s} } [D_{\phiv}(\xv_r, r)]  \big\|^2 \hspace{-0.12em}, \label{eq:consistency_loss}
\end{align}
where $r$ and $s$ are sampled from predefined distributions. Sampling from $p_{r|s}$ is implemented by solving \cref{eq:anderson_bwd} backward from $\xv_s$, replacing the score function with the network-estimated one $D_{\phiv}$ via \cref{eq:score_repara_in_denoiser}. For sampling from $p_s$, we first sample $\xv_\tau$ for $\tau > s$ and $\tau > \zeta$, then sample from $p_{s|\tau}$ in a manner analogous to sampling from $p_{r|s}$. 

It can be shown that if $D_{\phiv}$ minimizes the consistency loss for all $r, s$ and perfectly learns the score function for $t > \zeta$, then $\frac{D_{\phiv}(\xv_t, t) - \xv_t}{\sigma_t^2}$ becomes an exact estimator of the score function for all $t \in [0, T]$. Consequently, the distribution $p_0 = p_\text{data}$ can be sampled by solving \cref{eq:anderson_bwd}.
\vspace{-0.1em}

\citet{DarasDD2024} demonstrated the effectiveness of this framework only in fine-tuning latent diffusion models, leaving its efficacy when training from scratch unreported. Moreover, as sampling from $p_{r|s}$ depends on the model’s approximation of the score (which is particularly challenging to estimate accurately for $t < \zeta$) rather than the ground truth, there remains a gap between the theoretical framework and its practical implementation. This gap limits the extent to which the algorithm’s effectiveness is supported by their theoretical results.
\vspace{-0.5em}

\section{Theoretical Limit of Deconvolution}
\label{sec:deconv}
In this section, we evaluate the complexity of a deconvolution problem when the data corruption process is modelled using a forward diffusion process. Through the framework of deconvolution theory, we demonstrate that while \citet{DarasDD2024} showed that diffusion models can be trained using noisy samples, obtaining a sufficient number of samples to train high-quality models is practically infeasible.
\vspace{-0.25em}

The following two theorems establish that the optimal convergence rate for estimating the data density is $\Oc(\log n)^{-2}$. These results, derived using standard deconvolution theory~\citep{Meister2009} under a Gaussian noise assumption, highlight the inherent difficulty of the problem. We present the result for $d = 1$, which suffices to illustrate the challenge.  
\vspace{-0.7em}
\begin{restatable}{theorem}{MISEUpperBdound}
\label{thm:MISE_upper_bound}
Assume $ \Yc $ is generated according to \eqref{eq:gen_conv_samples} with $ \epsilon \sim \Nc(0, \sigma_\zeta^2) $ and $ p_\text{data} $  is a univariate distribution.  Under some weak assumptions on $ p_\text{data}$, for a sufficiently large sample size $ n $, there exists an estimator $ \hat{p}(\cdot; \mathcal{Y}) $ such that
    \begin{align}
        \mathrm{MISE}(\hat{p}, p_\textrm{data}) \leq C \cdot \frac{ \sigma_\zeta^4}{(\log n)^2},\label{eq:MISE_upper_bound}
    \end{align}
    where $ C $ is determined by $ p_\textrm{data} $.
\end{restatable}
\begin{restatable}{theorem}{MISELowerBound}
\label{thm:MISE_lower_bound}
In the same setting as \cref{thm:MISE_upper_bound}, for an arbitrary estimator $ \hat{p}(\cdot; \mathcal{Y}) $ of $p_\textrm{data}$ based on $\Yc$, 
\begin{align}
	\mathrm{MISE}(\hat{p}, p_\textrm{data}) \geq K \cdot (\log n)^{-2},
\end{align}
where $K>0$ is determined by $p_\textrm{data}$ and error distribution~$h$. 
\end{restatable}
The optimal convergence rate $\Oc(\log n)^{-2}$ indicates that reducing the MISE to one-fourth of its current value requires an additional $ n^2 - n $ samples. In contrast, under the error-free scenario, the optimal convergence rate is known to be $ \Oc(n^{-4/5})$ \citep{Wand1998}, where reducing the MISE to one-fourth of its current value would only necessitate approximately $ 4.657n $ additional samples.

The pessimistic rate indicates that effectively training a generative model using only corrupted samples with Gaussian noise is nearly impossible. Consequently, this implies that training from scratch,  using only noisy images, with the consistency loss discussed in \cref{prel:deconv_through_consist_const}, is infeasible. Notably, as indicated by \cref{eq:MISE_upper_bound}, this difficulty becomes significantly more severe with larger $\sigma_{\zeta}^2$, while a large $\sigma_{\zeta}^2$ is typically required to alter the original samples significantly to address copyright and privacy concerns.

To address the pessimistic statistical rate, we propose pretraining diffusion models on a small set of copyright-free samples. While this limited dataset can only capture a subset of the features and variations of the full true data distribution, we argue that it provides valuable prior information, enabling the model to start from a point much closer to the ground distribution compared to random weight initialization. For example, for image generation, pretraining allows the model to learn common features and structures shared among samples, such as continuity, smoothness, edges, and general appearance of typical object types.

Unfortunately, our empirical study in \cref{sec:emp} will show that the consistency loss-based method discussed in \cref{prel:deconv_through_consist_const} cannot deliver promising results even after pretraining. We suspect that this is caused by the gap between their theoretical framework and the practical implementation. As a result, we propose SFBD in \cref{sec:SFBD} to bridge such a gap.

\section{Stochastic Forward–Backward Deconvolution}
\label{sec:SFBD}
In this section, we introduce a novel method for solving the deconvolution problem that integrates seamlessly with the existing diffusion model framework. As our approach involves iteratively applying the forward diffusion process described in \cref{eq:fwd_diff}, followed by a backward step with an optimized drift, we refer to this method as Stochastic Forward-Backward Deconvolution (SFBD), as described in \cref{alg:SFBD}. 

The proposed algorithm begins with a small set of clean data, $\Dc_\text{clean}$, for pretraining, followed by iterative optimization using a large set of noisy samples. As demonstrated in \cref{sec:emp}, decent quality images can be achieved on datasets such as CIFAR-10 \citep{Krizhevsky2009} and CelebA \citep{LiuLWT2015} using as few as 50 clean images. During pretraining, the algorithm produces a neural network denoiser, $D_{{\phiv}_0}$, which serves as the initialization for the subsequent iterative optimization process. Specifically, the algorithm alternates between the following two steps: for $k = 1, 2, \ldots K$, 
\begin{enumerate}
	\item (Backward Sampling) This step can be intuitively seen as a denoising process for samples in $\Dc_\text{noisy}$ using the backward SDE \cref{eq:anderson_bwd}. In each iteration, we use the best estimation of the score function so far induced by $D_{{\phiv}_{k-1}}$ through \cref{eq:score_repara_in_denoiser}. 
	\item (Denoiser Update) Fine-tune denoiser $D_{{\phiv}_{k-1}}$ to obtain $D_{{\phiv}_k}$ by minimizing \cref{eq:denoiser_loss} with the denoised samples obtained in the previous step. 
\end{enumerate}
The following proposition shows that when $\Dc_\text{noisy}$ contains sufficiently many samples to characterize the true noisy distribution $p_{\rm data} \conv h$, when $K \rightarrow \infty$, the diffusion model implemented by denoiser $D_{{\phiv}_K}$ has the sample distribution converging to the true $p_\text{data}$. 


        


\begin{algorithm}[!t]
\DontPrintSemicolon
   \caption{Stochastic Forward–Backward Deconvolution. (Given sample set $\Dc$, $p_\Dc$ denotes the corresponding empirical distribution.)}
   \label{alg:SFBD}

    \KwIn{clean data: $\Dc_\text{clean} = \{\xv^{(i)}\}_{i=1}^M$, noisy data: $\Dc_\text{noisy} = \{\yv_\tau^{(i)}\}_{i=1}^N$, number of iterations: $K$.} 

    \tcp{Initialize Denoiser}
    
    $\phiv_0 \leftarrow$ Pretrain $D_{\phiv}$ using \cref{eq:denoiser_loss} with $p_0 = p_{\Dc_\text{clean}}$

    \For{$k = 1$ to $K$}{
        \tcp{Backward Sampling}
        
        $\Ec_k \leftarrow \{ \yv_0^{(i)} : \forall \yv_\tau^{(i)} \in \Dc_\text{noisy}$, solve backward SDE \cref{eq:anderson_bwd} from $\tau$ to $0$, starting from $\yv_\tau^{(i)}$, where the score function is estimated as $\frac{D_{\phiv_{k-1}}(\xv_t, t) - \xv_t}{\sigma_t^2} \}$
        
        \tcp{Denoiser Update}
        
        $\phiv_k \leftarrow$ Train $D_{\phiv}$ by minimizing \cref{eq:denoiser_loss} with $p_0 = p_{\Ec_k}$
    }

    \KwOut{Final denoiser $D_{{\phiv}_K}$}
\end{algorithm}

\begin{restatable}{proposition}{convSFBD}
\label{prop:conv_SFBD}
	Let $p^*_t$ be the density of $\xv_t$ obtained by solving the forward diffusion process \cref{eq:fwd_diff} with $\xv_0 \sim p_\text{data}$, where we have $p^*_\zeta = p_{\rm data} \conv h$. Consider a modified \cref{alg:SFBD}, where the empirical distribution $P_{\Dc_\text{noisy}}$ is replaced with the ground truth $p^*_\zeta$.  Correspondingly, $p_{{\Ec}_k}$ becomes $p_0^{(k)}$, the distribution of $\xv_0$ induced by solving:
	\begin{align}
		\diff \xv_t = - g(t)^2 \, \sv_{\phiv_{k-1}}(\xv_t, t) \diff t + g(t) \diff \bar\wv_t, ~\xv_\zeta \sim p^*_\zeta \label{eq:conv_SFBD:bwd}
	\end{align}
	from $\zeta$ to $0$, where $\sv_{\phiv_k}(\xv_t, t) = \frac{D_{\phiv_k}(\xv_t, t) - \xv_t}{\sigma_t^2}$, $g(t) = (\frac{\diff \sigma^2_t}{\diff t})^{\sfrac{1}{2}}$ and $D_{{\phiv}_k}$ is obtained by minimizing \eqref{eq:denoiser_loss} according to \cref{alg:SFBD}. Assume $D_{{\phiv}_k}$ reaches the optimal for all~$k$. Under mild assumptions, for $k \geq 0$, we have
	\begin{align}
		\KL{p_\text{data}}{p_0^{(k)}} \geq \KL{p_\text{data}}{p_0^{(k+1)}}. \label{eq:conv_SFBD:monotone}
	\end{align}
	In addition, for all $K \geq 1$ and $\uv \in \Rb^d$, we have
	\begin{align*}
			\min_{k = 1,\ldots K}\left|\Phi_{p_\text{data}}(\uv) - \Phi_{p_0^{(k)}}(\uv)\right|\leq \exp \big(\frac{\sigma_\zeta^2}{2} \|\uv\|^2 \big) \sqrt{\frac{2 M_0}{K}},
	\end{align*}	
	 where
	\begin{align*}
		M_0 = \tfrac{1}{2} \int_0^\zeta g(t)^2 \Eb_{p_{t}^*}\big\|\nabla \log p^*_t(\xv_t) - \sv_{\phiv_0}(\xv_t, t) \big\|^2 \diff t. 
	\end{align*}
\end{restatable}
\Cref{prop:conv_SFBD} shows that, after sufficiently many iterations of backward sampling and denoiser updates, the distribution of denoised samples produced by the backward sampling step converges to the true data distribution at a rate of $\Oc(1 / \sqrt{K})$ in terms of the characteristic function. This convergence implies that the corresponding densities and thus the distributions also become close. Consequently, fine-tuning the denoiser on these denoised samples during the Denoiser Update step enables the diffusion model to generate samples that approximately follow the true data distribution, thereby solving the deconvolution problem.

One might argue that the norm in the convergence bound depends not only on the iteration count $K$ but also on the norm of $\uv$, suggesting the existence of nontrivial approximation gaps when $|\uv|$ is large, regardless of how large $K$ is. We clarify that in practice, the contribution of the characteristic function at large $|\uv|$ is typically negligible. For distributions with smooth and bounded densities, characteristic functions decay rapidly, often at an exponential or super-polynomial rate. We elaborate further on this point in \cref{appx:proof:SFBD}, following the proof of the proposition.

Lastly, we note that this convergence result assumes access to infinitely many noisy samples and should be distinguished from the sample efficiency bounds discussed in \cref{sec:deconv}.

\textbf{The importance of pretraining.} \cref{prop:conv_SFBD} also highlights the critical role of pretraining, as it allows the algorithm to begin fine-tuning from a point much closer to the true data distribution. Specifically, effective pretraining ensures that $\sv_{\phiv_0}$ closely approximates the ground-truth score, leading to a smaller  $M_0$  in \cref{prop:conv_SFBD}. This, in turn, reduces the number of iterations $K$ required for the diffusion model to generate high-quality samples.
\vspace{0.245em}

\textbf{The practical limits of increasing $K$.}  While \cref{prop:conv_SFBD} suggests that increasing the number of iterations $K$ can continuously improve sample quality, practical limitations come into play. Sampling errors introduced during the backward sampling process, as well as imperfections in the denoiser updates, accumulate over time. These errors eventually offset the benefits of additional iterations, as demonstrated in \cref{sec:emp}. This observation further highlights the importance of pretraining to mitigate the impact of such errors and achieve high-quality samples with fewer iterations.
\vspace{0.245em}

\textbf{Alternative methods for backward sampling.} While the backward sampling in \cref{alg:SFBD} is presented as a naive solution to the backward SDE in \cref{eq:anderson_bwd}, the algorithm is not limited to this approach. Any backward SDE and solver yielding the same marginal distribution as \cref{eq:anderson_bwd} can be employed. Alternatives include PF-ODE, the predictor-corrector sampler \citep{SongDCKKEP2021}, DEIS \citep{ZhangC2023}, and the $2^\text{nd}$ order Heun method used in EDM \citep{KarrasAAL22}. Compared to the Euler–Maruyama method, these approaches require fewer network evaluations and offer improved error control for imperfect score estimation and step discretization. As the algorithm generates $\Ec_k$ that contains samples closer to $p_\text{data}$ with increasing $k$, clean images used for pretraining can be incorporated into $\Ec_k$ to accelerate this process. In our empirical study, this technique is applied whenever clean samples and noisy samples (prior to corruption) originate from the same distribution.

\begin{table}[!t]
\caption{Performance comparison of generative models. When $\sigma_\zeta > 0$, the models are trained on noisy images corrupted by Gaussian noise $\mathcal{N}(\zero, \sigma_\zeta^2 \Iv)$ after rescaling pixel values to $[-1, 1]$. For pretrained models, 50 clean images are randomly sampled from the training datasets for pretraining. \underline{Underscored} results are produced by this work. \textbf{Bolded} values indicate the best performance.}
\label{tb:performance_compare}
\resizebox{\columnwidth}{!}{%
\begin{tabular}{@{}lccr|ccr@{}}
\toprule
\multirow{2}{*}{Method} & \multicolumn{3}{c}{CIFAR10 (32 x 32)}                                          & \multicolumn{3}{c}{CelebA  (64 x 64)}                                              \\ \cmidrule(l){2-7} 
                        & $\sigma_\zeta$ & Pretrain             & FID    & $\sigma_\zeta$ & Pretrain & FID \\ \midrule
DDPM \citep{HoJA2020}                   & 0.0   & No                   & 4.04   & 0.0                       & No                           & \textbf{3.26}                    \\
DDIM \citep{SongME2021}                   & 0.0   & No                   & 4.16   & 0.0                       & No                           & 6.53                    \\
EDM \citep{KarrasAAL22}                    & 0.0   & No                   & \textbf{1.97}   & -                       & -                           & -                       \\ \midrule
SURE-Score \citep{AaliAKT2023}             & 0.2                       & Yes                  & 132.61 & -                         & -                            & -                       \\
EMDiff \citep{BaiWCS2024}            & 0.2                       & Yes                  & 86.47  & -                         & -                            & -                       \\
TweedieDiff \citep{DarasDD2024}       & 0.2                       & No                   & \underline{167.23} & 0.2                       & No                           & \underline{246.95}                  \\
TweedieDiff \citep{DarasDD2024}       & 0.2                       & Yes                  & \underline{65.21}  & 0.2                       & Yes                          & \underline{58.52}                   \\
TweedieDiff+ \citep{DarasCD25}      & 0.2                       & Yes                  & 8.05  & 0.2                       & Yes                          & -                   \\
SFBD (Ours)             & 0.2                       & Yes                  & \underline{\textbf{13.53}} & 0.2                       & Yes                          &      \underline{\textbf{6.49}}                   \\
\bottomrule
\end{tabular}
}
\end{table}
\begin{figure}[t]
\centering
\includegraphics[width=0.8\columnwidth]{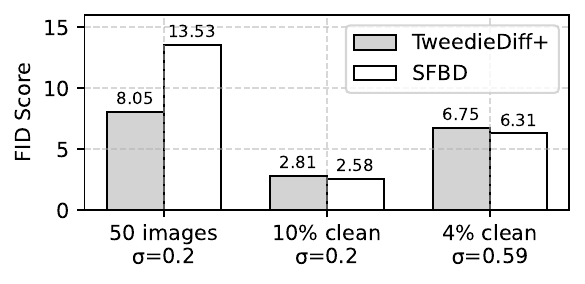}	
\caption{TweedieDiff+ vs SFBD on CIFAR10}
\label{fig:TweedieDiffPlus_vs_SFBD}
\end{figure}

\textbf{Relationship to the consistency loss.} SFBD can be seen as an algorithm that enforces the consistency constraint across all positive time steps and time zero. Specifically, we have
\vspace{0.1em}
\begin{restatable}{proposition}{RELCONSSFBD}
\label{prop:rel_btw_cons_SFBD}
Assume that the denoising network $D_{\phiv}$ is implemented to satisfy $D_{\phiv}(\cdot, 0) = \textrm{Id}(\cdot)$. When $r = 0$, the consistency loss in \cref{eq:consistency_loss} is equivalent to the denoising noise in \cref{eq:denoiser_loss} for $t = s$. 
\end{restatable}
\vspace{0.1em}

The requirement that $D_{\phiv}(\cdot, 0) = \textrm{Id}(\cdot)$ is both natural and intuitive, as $D_{\phiv}(\xv_0, 0)$ approximates $\Eb[\xv_0 | \xv_0] = \xv_0$. This fact is explicitly enforced in the design of the EDM framework \citep{KarrasAAL22}, which has been widely adopted in subsequent research.

A key distinction between SFBD and the original consistency loss implementation is that SFBD does not require sampling from $p_{r|s}$ or access to the ground-truth score function induced by the unknown data distribution $p_\text{data}$. This is because, in the original implementation, $p_0 = p_\text{data}$, whereas in SFBD,  $p_0 = p_0^{(k)}$, as defined in \cref{prop:conv_SFBD}, and is obtained iteratively through the backward sampling step. As $k$  increases,  $p_0^{(k)}$  converges to  $p_\text{data}$, ensuring that the same consistency constraints are eventually enforced. Consequently, SFBD bridges the gap between theoretical formulation and practical implementation that exists in the original consistency loss framework.


%

\section{Empirical Study}
\label{sec:emp}
In this section, we demonstrate the effectiveness of the SFBD framework proposed in \cref{sec:SFBD}. Compared to other models trained on noisy datasets, SFBD consistently achieves superior performance across all benchmark settings. Additionally, we conduct ablation studies to validate our theoretical findings and offer practical insights for applying SFBD effectively.

\begin{figure}[!t]
    \centering
    \setlength{\tabcolsep}{1pt} 
    \setlength{\fboxrule}{1pt} 
    \def\imagewidth{0.15\linewidth} 
    \def\imagewidthb{0.19\linewidth} 
    \def\celebaimgidxa{000000.png} 
    \def\celebaimgidxb{000006.png} 
    \renewcommand{\arraystretch}{0.5}
    \centering
    \resizebox{\columnwidth}{!}{
    \begin{tabular}{c}  
        \begin{tabular}{cccccccc}
            \tiny{\makecell[c]{Noisy\\Observ.}} &
            \tiny{\makecell[c]{SURE-\\Score}} &
            \tiny{\makecell[c]{Amb-Diff}} &
            \tiny{\makecell[c]{EMDiff}} &
            \tiny{\makecell[c]{Tweedie\\ Diff w/o PT}} &
            \tiny{\makecell[c]{Tweedie\\ Diff w/ PT}} &
            \tiny{\makecell[c]{SFBD \\ (\textbf{Ours})}} &
            \tiny{\makecell[c]{Ground\\Truth}} \\
            \begin{overpic}[width=\imagewidth]{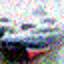}\end{overpic} & 
            \begin{overpic}[width=\imagewidth]{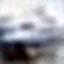}\end{overpic} & 
            \begin{overpic}[width=\imagewidth]{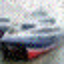}\end{overpic} & 
            \begin{overpic}[width=\imagewidth]{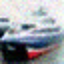}\end{overpic} & 
            \begin{overpic}[width=\imagewidth]{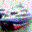}\end{overpic} & 
            \begin{overpic}[width=\imagewidth]{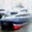}\end{overpic} & 
            \begin{overpic}[width=\imagewidth]{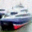}\end{overpic} & 
            \begin{overpic}[width=\imagewidth]{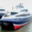}\end{overpic} \\ 

            \begin{overpic}[width=\imagewidth]{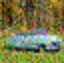}\end{overpic} &
            \begin{overpic}[width=\imagewidth]{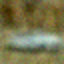}\end{overpic} &
            \begin{overpic}[width=\imagewidth]{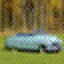}\end{overpic} &
            \begin{overpic}[width=\imagewidth]{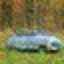}\end{overpic} &
            \begin{overpic}[width=\imagewidth]{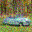}\end{overpic} & 
            \begin{overpic}[width=\imagewidth]{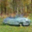}\end{overpic} & 
            \begin{overpic}[width=\imagewidth]{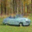}\end{overpic} & 
            \begin{overpic}[width=\imagewidth]{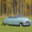}\end{overpic} \\

            \begin{overpic}[width=\imagewidth]{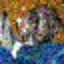}\end{overpic} &
            \begin{overpic}[width=\imagewidth]{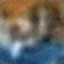}\end{overpic} &
            \begin{overpic}[width=\imagewidth]{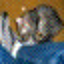}\end{overpic} &
            \begin{overpic}[width=\imagewidth]{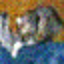}\end{overpic} &
            \begin{overpic}[width=\imagewidth]{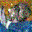}\end{overpic} & 
            \begin{overpic}[width=\imagewidth]{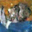}\end{overpic} & 
            \begin{overpic}[width=\imagewidth]{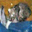}\end{overpic} & 
            \begin{overpic}[width=\imagewidth]{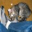}\end{overpic} \\
        \end{tabular}
    \end{tabular}
    }

~

~

    \resizebox{\columnwidth}{!}{
    \begin{tabular}{c}
        \begin{tabular}{cccccccc}
            \tiny{\makecell[c]{Noisy Observ.}} &
            \tiny{\makecell[c]{Tweedie\\ Diff w/o PT}} &
            \tiny{\makecell[c]{Tweedie\\ Diff w/ PT}} &
            \tiny{\makecell[c]{SFBD  (\textbf{Ours})}} &
            \tiny{\makecell[c]{Ground Truth}} \\
            \begin{overpic}[width=\imagewidthb]{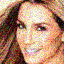}\end{overpic} & 
            \begin{overpic}[width=\imagewidthb]{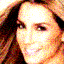}\end{overpic} & 
            \begin{overpic}[width=\imagewidthb]{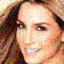}\end{overpic} & 
            \begin{overpic}[width=\imagewidthb]{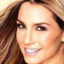}\end{overpic} & 
            \begin{overpic}[width=\imagewidthb]{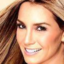}\end{overpic} \\ 


            \begin{overpic}[width=\imagewidthb]{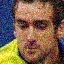}\end{overpic} & 
            \begin{overpic}[width=\imagewidthb]{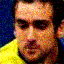}\end{overpic} & 
            \begin{overpic}[width=\imagewidthb]{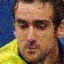}\end{overpic} & 
            \begin{overpic}[width=\imagewidthb]{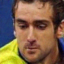}\end{overpic} & 
            \begin{overpic}[width=\imagewidthb]{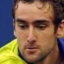}\end{overpic} \\ 

        \end{tabular}
    \end{tabular}
    }

    \caption{Denoised samples of CIFAR-10 (up) and CelebA (down). (Noise level $\sigma_\zeta = 0.2$)} 

    \label{fig:model_benchmark_visual}
\end{figure}

\textbf{Datasets and evaluation metrics.} The experiments are conducted on the CIFAR-10 \citep{Krizhevsky2009} and CelebA \citep{LiuGL2022} datasets, with resolutions of  $32 \times 32$  and  $64 \times 64$, respectively. CIFAR-10 consists of 50,000 training images and 10,000 test images across 10 classes. CelebA, a dataset of human face images, includes a predefined split of 162,770 training images, 19,867 validation images, and 19,962 test images. For CelebA,  images were obtained using the preprocessing tool provided in the DDIM official repository  \citep{SongME2021}.
\begin{figure*}
\centering     
\subfigure[Clean Image Ratio]{\label{fig:ablation_cifar10:clean_ratio}\includegraphics[width=0.29\textwidth]{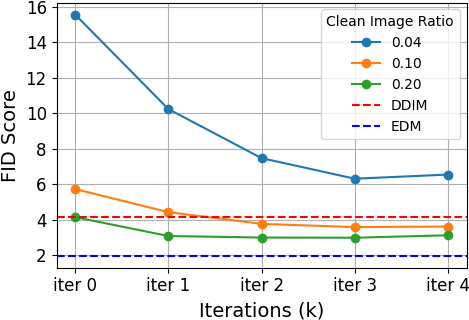}}\hfill
\subfigure[Noise Level]{\label{fig:ablation_cifar10:sigma}\includegraphics[width=0.29\textwidth]{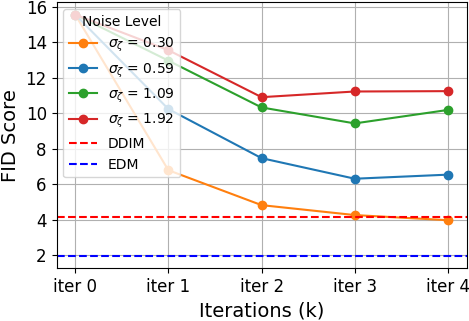}}\hfill
\subfigure[Pretraining on Similar Datasets]{\label{fig:ablation_cifar10:pretrain}\includegraphics[width=0.29\textwidth]{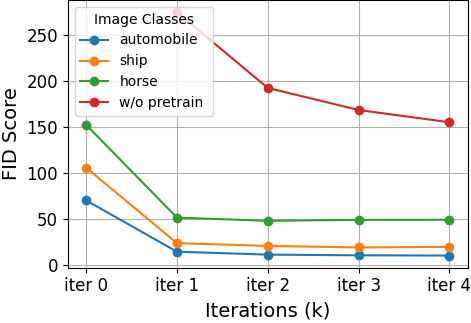}}
\vspace{-0.9em}
\caption{SFBD performance on CIFAR-10 under various conditions. Unless specified, the clean image ratio is $0.04$ and the noise level $\sigma_\zeta$ is $0.59$. In (a) and (b), FID at iteration 0 corresponds to the pretrained model. In (c), models are pretrained on clean images from the ``truck'' class, with FID at iteration 0 measuring the distance between these clean images and those used for fine-tuning. For the w/o pretraining setting, models are trained on the full CIFAR-10 dataset with $\sigma_\zeta = 0.59$.}
\label{fig:ablation_cifar10}
\vspace{-1em}
\end{figure*}

We evaluate image quality using the Frechet Inception Distance (FID), computed between the reference dataset and 50,000 images generated by the models. Generated samples for FID computation are presented in \cref{appx:sample_results}.

\begin{figure}
	\centering
	\includegraphics[width=1\columnwidth]{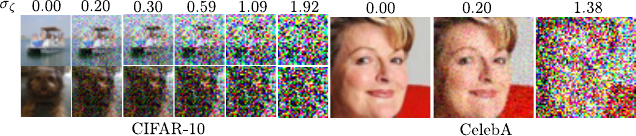}

	\caption{Noisy images with different $\sigma_\zeta$.}
	\label{fig:noisy_img_var_sigma}

\end{figure}

\textbf{Models and other configurations.} We implemented SFBD algorithms using the architectures proposed in EDM \citep{KarrasAAL22} as well as the optimizers and hyperparameter configurations therein. All models are implemented in an unconditional setting, and we also enabled the non-leaky augmentation technique \citep{KarrasAAL22} to alleviate the overfitting problem. For the backward sampling step in SFBD, we adopt the $2^\text{nd}$-order Heun method \citep{KarrasAAL22}. More information is provided in \cref{appx:expConfig}.

\vspace{0.25em}

\subsection{Performance Comparison}
\vspace{0.25em}

In \cref{tb:performance_compare}, we compare SFBD with representative models for training on noisy images. SURE-Score \citep{AaliAKT2023} and EMDiff \citep{BaiWCS2024} tackle general inverse problems using Stein’s unbiased risk estimate and expectation-maximization, respectively. TweedieDiffusion \citep{DarasDD2024} applies the original consistency loss from \cref{eq:consistency_loss}, while \citet{DarasCD25} introduce TweedieDiff+ with a simplified implementation that improves performance.
\vspace{0.2em}

Following the experimental setup of \citet{BaiWCS2024}, images are corrupted by adding independent Gaussian noise with a standard deviation of  $\sigma_\zeta = 0.2$  to each pixel after rescaling pixel values to $[-1, 1]$. For reference, we also include results for models trained on clean images ($\sigma_\zeta = 0$). In cases with pretraining, the models are initially trained on 50 clean images randomly sampled from the training datasets. For all results presented in this work, the same set of 50 sampled images is used.
\vspace{0.2em}

As shown in \cref{tb:performance_compare}, SFBD consistently produces higher-quality images than all baselines except TweedieDiff+, with further visual evidence provided in \cref{fig:model_benchmark_visual}, where denoised samples are generated by evaluating the backward SDE from noisy training images. Notably, on CelebA, SFBD achieves performance comparable to DDIM trained entirely on clean data. While TweedieDiff benefits from pretraining, its performance remains inferior to SFBD. In fact, we observe that the original consistency loss \eqref{eq:consistency_loss} yields only limited improvement post-pretraining, with FID scores deteriorating soon after the loss is applied. TweedieDiff+, which adopts a simplified version of the consistency loss, outperforms SFBD when only a very limited amount of clean data is available, likely due to overfitting during our model’s pretraining. However, this effect diminishes as more clean data is introduced, allowing SFBD to surpass TweedieDiff+, as shown in \cref{fig:TweedieDiffPlus_vs_SFBD}. We attribute this to SFBD’s more stable fine-tuning via the score-matching loss, which avoids the additional constraints imposed by consistency-based methods.

\begin{figure*}
	\centering
	\includegraphics[width=0.5\columnwidth]{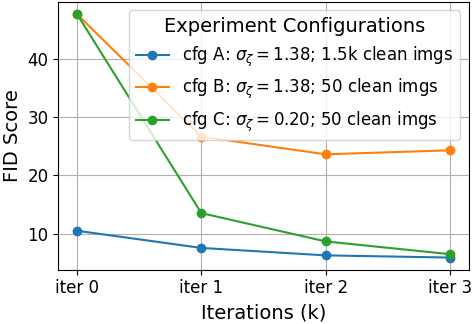}~~~~
	\raisebox{-0.1em}{\includegraphics[width=1.4\columnwidth]{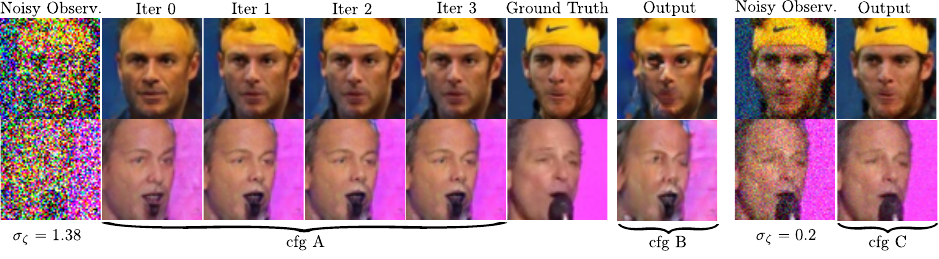}}
	\caption{(Left) SFBD performance on CelebA under three configurations, with FID at iteration 0 for the pretrained model. (Right) Denoised samples generated by the backward SDE, starting from a noisy image in the training dataset. For cfg A, results are shown after each fine-tuning iteration, while cfg B and cfg C are shown at their minimum FID iterations.}
	\label{fig:var_config_celeba}
\vspace{-1.2em}
\end{figure*}

\subsection{Ablation Study}
\label{sec:exp:ablation}
In this section, we investigate how SFBD's performance varies with clean image ratios, noise levels, and pretraining on similar datasets. The results align with our discussion in \cref{sec:deconv} and \cref{sec:SFBD} and provide practical insights. Experiments are conducted on CIFAR-10, with the default $\sigma_\zeta = 0.59$. This noise level significantly alters the original images, aligning with our original motivation to address potential copyright concerns (see \cref{fig:noisy_img_var_sigma}). 

\textbf{Clean image ratio.} \cref{fig:ablation_cifar10:clean_ratio} shows the FID trajectories across fine-tuning iterations $k$ for different clean image ratios. With just 4\% clean images, SFBD achieves strong performance (FID: 6.31) and outperforms DDIM with 10\% clean images. While higher clean image ratios further improve performance, the gains diminish as a small amount of clean data already provides sufficient high-frequency features (e.g., edges and local details) to capture feature variations. Since these features are shared across images, additional clean data offers limited improvement.

These findings suggest that practitioners with limited clean datasets should focus on collecting more copyright-free data to enhance performance. Notably, when clean images are scarce, the marginal gains from additional fine-tuning iterations $k$ are greater than when more clean data is available. Therefore, in scenarios where acquiring clean data is challenging, increasing fine-tuning iterations can be an effective alternative to improve results.

\textbf{Noise level.} \cref{fig:ablation_cifar10:sigma} shows SFBD's sampling performance across fine-tuning iterations for different noise levels, using the values from $2^\text{nd}$ order Heun sampling in EDM \citep{KarrasAAL22}. The impact of noise on the original images is visualized in \cref{fig:noisy_img_var_sigma}. As shown in \cref{fig:ablation_cifar10:sigma}, increasing $\sigma_\zeta$ significantly degrades SFBD's performance. This is expected, as higher noise levels obscure more features in the original images. Furthermore, as suggested by \cref{thm:MISE_upper_bound}, higher $\sigma_\zeta$ demands substantially more noisy images, which cannot be compensated by pretraining on a small clean image set. Importantly, this performance drop is a mathematical limitation discussed in \cref{sec:deconv}, rather than an issue solvable by better deconvolution algorithms. In \cref{sec:exp:futher_discuss}, we show that slightly increasing pretraining clean image set can yield strong results, even at reasonably high noise levels on CelebA.

\textbf{Pretraining with clean images from similar datasets.} \cref{fig:ablation_cifar10:pretrain} evaluates SFBD's performance when fine-tuning on image sets from different classes, with the model initially pretrained on clean truck images. The results show that the closer the noisy dataset is to the truck dataset (as indicated by the FID at iter 0), the better the model performs after fine-tuning. This is expected, as similar datasets share common features that facilitate learning the target data distribution.
Interestingly, even when the pretraining dataset differs significantly from the noisy dataset, the model still outperforms the version without pretraining. This is because unrelated datasets often share fundamental features, such as edges and local structures.  \textit{Therefore, practitioners should always consider pretraining before fine-tuning on target noisy datasets, while more similar pretraining datasets yield better final sampling performance.}

\begin{table}[!t]
\caption{Comparison of SFBD and Restormer \citep{ZamirAKHKY2022} on denoising tasks.}\label{tb:SFBD_vs_Restormer}
\vspace{-0.2em}
\resizebox{\columnwidth}{!}{%
\begin{tabular}{@{}llcccc|llcccc@{}}
\toprule
\multicolumn{6}{c|}{{\large {CIFAR-10 (4\% clean images)}}} & \multicolumn{6}{c}{\large {CelebA}} \\
\midrule
$\sigma$ & Model & Iter 1 & Iter 2 & Iter 3 & Iter 4 
         & Setting & Model & Iter 1 & Iter 2 & Iter 3 & Iter 4 \\
\midrule
0.30 & SFBD      & 6.16  & 3.42  & 2.68  & 2.35 
     & 50 clean imgs & SFBD      & 47.69 & 10.05 & 5.63 & 3.93 \\
     & Restormer & \multicolumn{4}{c|}{53.87}
     & $\sigma=0.2$   & Restormer & \multicolumn{4}{c}{18.90} \\
\midrule
0.59 & SFBD      & 10.23 & 7.47  & 6.31  & 6.54 
     & 1.5k clean imgs & SFBD      & 9.05  & 5.76  & 4.56 & 3.98 \\
     & Restormer & \multicolumn{4}{c|}{99.99}
     & $\sigma=1.38$   & Restormer & \multicolumn{4}{c}{227.91} \\
\midrule
1.09 & SFBD      & 12.68 & 9.39  & 9.08  & 10.14 
     & \multicolumn{6}{c}{\raisebox{-0.8em}{\textemdash}} \\
     & Restormer & \multicolumn{4}{c|}{132.69} 
     & \multicolumn{6}{c}{} \\
\bottomrule
\end{tabular}
}
\vspace{-1em}
\end{table}

\subsection{Further Discussions}
\label{sec:exp:futher_discuss}

\textbf{Additional results on CelebA.} \cref{fig:var_config_celeba} presents SFBD performance trajectories on CelebA under three configurations. While \cref{tb:performance_compare} reports results using configuration (cfg) C to align with benchmarks, this setup is impractical due to its low noise level, which fails to address copyright and privacy concerns. As illustrated in \cref{fig:var_config_celeba} (right), the low noise level allows human observers to identify individuals and recover image details, with model-denoised images nearly identical to the originals. To address this, we report results for cfgs A and B with $\sigma_\zeta = 1.38$, concealing most original image information. While pretraining on 50 clean images performs poorly, increasing the size to 1.5k (still $< 1\%$ of the training dataset) achieves impressive results. At iteration 3, the model reaches FID 5.91, outperforming DDIM trained on clean images. This supports our discussion in \cref{sec:exp:ablation}: collecting more clean data significantly boosts performance when the clean dataset is small.

\vspace{0.25em}

\textbf{Features learned from noisy images.} As shown in \cref{fig:var_config_celeba}, when $\sigma_\zeta = 1.38$, almost all information from the original images is obscured, prompting the question: can the model learn from such noisy inputs, and how does this happen? In \cref{fig:var_config_celeba}, we plot the model’s denoised outputs in cfg A after each fine-tuning iteration. These outputs serve as samples for the next iteration, revealing what the model learns and adapts to in the process. For the first row, the pretrained model (iter 0) produces a face very different from the original, failing to recover features like a headband. This occurs because the clean dataset for pretraining lacks similar faces with headbands. Instead of random guesses, the model combines local features (e.g., face shapes, eyes) learned from the clean data with the global structure from the noisy images. This process combines previously learned features in new ways, helps the model better generalize, and gradually improves its ability to approximate the true distribution, as supported by \cref{prop:conv_SFBD}. Similarly, in the second row, the model learns to attach a goatee to the face despite the corresponding region in the original image being a microphone.

\vspace{0.25em}

\textbf{How SFBD differs from standard denoising algorithms.}
While SFBD is intuitively described as alternating between denoising and fine-tuning, it does not function like a traditional denoiser that reconstructs exact clean samples. Instead, it learns to match the full data distribution, allowing greater flexibility and producing more realistic outputs. In \cref{tb:SFBD_vs_Restormer}, we compare the FID of SFBD-denoised samples at each iteration ($\Ec_k$ in \cref{alg:SFBD}) against those denoised by Restormer~\citep{ZamirAKHKY2022}, a strong off-the-shelf denoiser. SFBD consistently yields significantly lower FIDs even after the first iteration, and the sample quality improves steadily with more updates.
\vspace{-0.25em}

These results also caution against replacing SFBD’s denoising process with a classical denoiser. Since models trained on denoised samples cannot surpass their targets in FID, the values in \cref{tb:SFBD_vs_Restormer} represent upper bounds. Notably, final SFBD models achieve lower generative FIDs (\cref{fig:ablation_cifar10:sigma}) than those from Restormer-denoised data, making it unlikely that a competitive model could be trained on such samples.
\vspace{-0.25em}

\textbf{Data leakage and sample memorization.} 
We note that SFBD is not intended to prevent leakage of the clean samples used for pretraining. As these samples are assumed to be public and copyright-free, leakage from this subset is not a concern. Instead, SFBD is specifically designed to protect \textit{sensitive} data. By construction, the model accesses only a single corrupted version of each sensitive sample during the entire training process. This design inherently limits the model’s ability to reconstruct copyrighted or private content, making SFBD privacy-preserving by nature. In \cref{appx:data_leakage}, we adopt the methodology of \citet{DarasDD2024} to evaluate privacy risks: we plot similarity score distributions and identify the sensitive sample most similar to any generated output. The results confirm that SFBD does not reconstruct sensitive data, supporting our privacy claims.
\vspace{-0.5em}


\section{Conclusion}
\label{sec:disc}
\vspace{-0.3em}
In this paper, we presented SFBD, a new deconvolution method based on diffusion models. Under mild assumptions, we theoretically showed that our method could guide diffusion models to learn the true data distribution through training on noisy samples. The empirical study corroborates our claims and shows that our model consistently achieves state-of-the-art performance in some benchmark tasks.

\clearpage 
\section*{Acknowledgement}
We gratefully acknowledge funding support from NSERC, the Canada CIFAR AI Chairs program and the Ontario Early Researcher program. Resources used in preparing this research were provided, in part, by the Province of Ontario, the Government of Canada through CIFAR, and companies sponsoring the Vector Institute.

\section*{Impact Statement}
This paper introduces SFBD, a framework for effectively training diffusion models primarily using noisy samples. Our approach enables data sharing for generative model training while safeguarding sensitive information.

For organizations utilizing personal or copyrighted data to train their models, SFBD offers a practical solution to mitigate copyright concerns, as the model never directly accesses the original samples. This mathematically guaranteed framework can promote data-sharing by providing a secure and privacy-preserving training method.

However, improper implementation poses a risk of sensitive information leakage. A false sense of security could further exacerbate this issue, underscoring the importance of rigorous validation and responsible deployment.

\bibliography{example_paper}
\bibliographystyle{icml2025}

\newpage
\appendix
\onecolumn

\section{A Brief Introduction to the Density Convolutions}
\label{appx:intro:density_deconv}
In this section, we give a brief discussion on the density convolution and how it is related to our problem. 

For simplicity, we stick to the case when $d = 1$. Consider the data generation process in \cref{eq:gen_conv_samples}. Let $p_y$ denote the density of the distribution of the noisy samples $y^{(i)}$. Then we have 
\begin{fact}
\label{fact:conv_density}
For $\omega \in \Rb$, 
\begin{align}
	p_y(\omega) = \int p_\text{data}(x) \; h(\omega-x) \diff x = (p_\text{data} \conv h) (\omega).
\end{align} 
\end{fact}
\begin{proof}
	This is because, for all measurable function $\psi$, we have
\begin{align*}
	\int \psi(\omega) p_y(\omega) \diff \omega &= \int \int \psi(x+\epsilon) \; p_\text{data}(x) h(\epsilon) \diff x \diff \epsilon = \int \int \psi(\omega) p_\text{data}(x) h(\omega-x) \diff x \diff w\\
	&= \int \psi(w) \left[\int p_\text{data}(x) \; h(\omega - x) \diff x \right]\diff \omega. 
\end{align*}
As the equality holds for all $\psi$, we have
\(
	p_y(\omega) = \int p_\text{data}(x) \; h(\omega-x) \diff x = (p_\text{data} \conv h) (\omega).
\)
\end{proof}
As a result, according to \cref{fact:conv_density}, the density convolution is naturally involved in our setting.

\begin{figure}[h]
    \centering
    \includegraphics[width=0.33\linewidth]{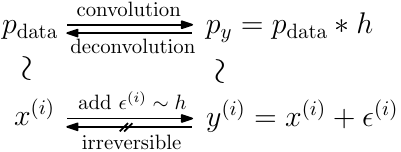}
    \caption{While the corruption process is irreversible at the sample level, a bijective relationship exists between the clean and noisy data distributions.}
    \label{fig:diagram_dist_conv}
\end{figure}

Then, we provide an alternative way to show why we can recover $p_\text{data}$ given $p_y$ and $h$. (Namely, we need to deconvolute $p_y$ to obtain  $p_\text{data}$.) Our discussion can be seen a complement of the discussion following  \cref{prop:conv_identify}. Let $\phi_p$ denote the characteristic function of the random variable with distribution $p$ such that 
\begin{align}
    \phi_p(t) = \int \exp(i t \omega) \; p(\omega) \diff \omega.
\end{align}
We note that the characteristic function of a density $p$ is its Fourier transform. As a result, through the dual relationship of multiplication and convolution under Fourier transformation \citep[Lemma A.5]{Meister2009}, we have 
\begin{align}
    \phi_{p_y}(t) = \phi_{p_\text{data}}(t) \; \phi_h (t).
\end{align}
As a result, given noisy data distribution $p_y$ and noise distribution $h$, we have
\begin{align}
    \phi_{p_\text{data}} (t) = \frac{\phi_{p_y}(t)}{\phi_h(t)}. 
\end{align}
Finally, we can recover $p_\text{data}$ through an inverse Fourier transform:
\begin{align}
    p_\text{data}(x) = (2\pi)^{-1} \int \exp(-itx) \; \phi_{p_\text{data}} (t) \diff t = (2\pi)^{-1} \int \exp(-itx) \; \frac{\phi_{p_y}(t)}{\phi_h(t)} \diff t.
\end{align}
We conclude this section by summarizing the relationship between data and noisy sample distributions in \cref{fig:diagram_dist_conv}.

\section{Proofs Related to Deconvolution Theory}
\label{appx:proof:deconv}
We first show the result suggesting it is possible to identify a distribution through its noisy version obtained by corrupting its samples by injecting independent Gaussian noises. 
\PROPCONVINDENTIFY*

\begin{lemma}
\label{appx:lem:kl_bd_char}
	Given two distributions $p$ and $q$ on $\Rb^d$. Let $\Phi_p(\uv)$ and $\Phi_q(\uv)$ be their characteristic functions. Then for all $\uv \in \Rb^d$, we have
	\begin{align}
			\bigl|\Phi_p(\uv) - \Phi_q(\uv)\bigr|
			\;\le\;
			\sqrt{2 \,D_{\mathrm{KL}}(p\;\|\;q)}.
	\end{align}
\end{lemma}

\begin{proof}
We note that 
\[
\Phi_p(\uv) \;=\; \mathbb{E}_p[\exp(i\uv^\top \xv)],
\quad
\Phi_q(\uv) \;=\; \mathbb{E}_q[\exp(i\uv^\top \xv)].
\]
Then for any $\uv \in \Rb^d$, we have
\begin{align*}
\bigl|\Phi_p(\uv) - \Phi_q(\uv)\bigr|
& \le
\left|\int_{\Rb^d} \exp(i\uv^\top\xv) p(\xv) \diff \xv - \int_{\Rb^d} \exp(i\uv^\top\xv) q(\xv) \diff \xv \right| \\
& = \left|\int_{\Rb^d} \exp(i\uv^\top\xv) \Big(p(\xv) -  q(\xv) \Big) \diff \xv \right|  \leq \int_{\Rb^d} \underbrace{\left|\exp(i\uv^\top\xv)\right|}_{=1}  \left| p(\xv)  -  q(\xv) \right| \diff \xv \\
& = \int_{\Rb^d}  \left| p(\xv)  -  q(\xv) \right| \diff \xv \\
& = 2 \; \|p - q\|_{\mathrm{TV}},
\end{align*}
where the last equality is due to Scheffe's theorem \citep[Lemma 2.1, p. 84]{Tsybakov09}).

Then, by Pinsker's inequality \citep[Lemma 2.5, p. 88]{Tsybakov09}, we have 
\[
\bigl|\Phi_p(\uv) - \Phi_q(\uv)\bigr|
\;\le\;
2 \; \|p - q\|_{\mathrm{TV}}
\;\le\;
\sqrt{2\,D_{\mathrm{KL}}(P\;\|\;Q)}.
\]
which completes the proof. 
\end{proof}
\begin{proof}[Proof of \cref{prop:conv_identify}]
	Note that, by the convolution theorem \citep[A.4]{Meister2009},  for all $\uv \in \Rb^d$, we have 
	\begin{align*}
		\Phi_{p * h}(\uv) = \Phi_{p}(\uv) \; \Phi_{h}(\uv) = \Phi_{p}(\uv) \; \exp \big(-\frac{\sigma_\zeta^2}{2} \|\uv\|^2 \big),
	\end{align*}
	as $h\sim \Nc(\mathbf{0}, \sigma_\zeta^2\mathbf{I})$ having $\Phi_h(\uv) = \exp \big(-\frac{\sigma_\zeta^2}{2} \|\uv\|^2 \big)$. Applying \cref{appx:lem:kl_bd_char}, we have
	\begin{align}
		\exp \big(-\frac{\sigma_\zeta^2}{2}\|\uv\|^2\big) \; \Big|\Phi_{p}(\uv) - \Phi_{q}(\uv)\Big|  = \big|\Phi_{p * h}(\uv) - \Phi_{q * h}(\uv) \big| \leq \sqrt{2\,D_{\mathrm{KL}}(p \conv h\|q\conv h)}. 
	\end{align}	
	Rearranging the inequality completes the proof. 
\end{proof}

We then derive the proofs regarding the sample complexity of the deconvolution problem. 
\MISEUpperBdound*
\begin{proof}
    The result is constructed based on the work by \citet{StefanskiC1990}. In particular, assuming that $p_\text{data}$ is continuous, bounded and has two bounded integrable derivatives such that 
    \begin{align}
        \int p''_\text{data} (x) \diff x < \infty,
    \end{align}
    we can construct a kernel based estimator of $p_\text{data}$ of rate
    \begin{align}
        \frac{\lambda^4}{4} \mu^2_{K, 2} \int p''_\text{data} (x) \diff x,
    \end{align}
    where $\mu^2_{\kappa, 2}$ is a constant determined by the selected kernel $\kappa$ and $\lambda$ is a function of number of samples $n$ gradually decreasing to zero as $n \rightarrow \infty$. It is required that $\lambda$ satisfies
    \begin{align}
        \frac{1}{2 \pi n \lambda} \exp( \frac{B^2 \sigma_\zeta^2}{\lambda^2}) \rightarrow 0 \label{appx:eq:mise_upper_bound:1}
    \end{align}
    as $n \rightarrow \infty$, where $B > 0$ is a constant depending on the picked kernel $\kappa$. Here, we assume we picked a kernel with $B < 1$.    
    
    To satisfy the constraint, we choose $\lambda (n) = \frac{\sigma_\zeta}{\sqrt{\log n}}$. Plugging it into \cref{appx:eq:mise_upper_bound:1}, we have 
    \begin{align}
        \lim_{n \rightarrow \infty} \frac{1}{n \lambda} \exp( \frac{B^2 \sigma_\zeta^2}{\lambda^2}) = \lim_{n \rightarrow \infty} \frac{\sqrt{\log n}}{n {\sigma_\zeta }} \exp{(B^2 \log n)} = \lim_{n \rightarrow \infty} \frac{\sqrt{\log n}}{n^{1-B^2} {\sigma_\zeta }}. 
    \end{align}
    To show $\lim_{n \rightarrow \infty} \frac{\sqrt{\log n}}{n^{1-B^2} {\sigma_\zeta }} = 0$, it suffices to show 
    $
        \lim_{n \rightarrow \infty} \frac{\log n}{n^{2 - 2 B^2} {\sigma_\zeta^2 }} = 0
    $. By L'Hopital's rule, we have
    \begin{align}
        \lim_{n \rightarrow \infty} \frac{\log n}{n^{2 - 2 B^2} {\sigma_\zeta^2 }} =  \lim_{n \rightarrow \infty} \frac{1}{(2 - 2 B^2) n^{2 - 2 B^2} {\sigma_\zeta^2 }} = 0
    \end{align}
    As a result, $\lambda (n) = \frac{\sigma_\zeta}{\sqrt{\log n}}$ is a valid choice, which gives the convergence rate
    $\frac{\sigma_\zeta^4}{(\log n)^2}$.
\end{proof}

\MISELowerBound*
\begin{proof}
    This result is a special case of Theorem 2.14 (b) in \citep{Meister2009}. When the error density is Gaussian, we have $\gamma = 2$. In addition, in the proof of \cref{thm:MISE_upper_bound}, we assumed that $p_\text{data}$ has two bounded integrable derivatives, which equivalently assumes $p_\text{data}$ satisfies the Soblev condition with smoothness degree $\beta = 2$ (see Eq. A.8,  \citealt{Meister2009}). Then the theorem shows $\mathrm{MISE}(\hat{p}, p_\textrm{data}) \geq \textrm{const} \cdot (\log n)^{-2\beta / \gamma} = \textrm{const} \cdot (\log n)^{-2}$. 
\end{proof}

\section{Proofs Related to the Results of SFBD}
\label{appx:proof:SFBD}
\def\bwdM{\overleftarrow{Q}_{0:\zeta}}
\def\fwdM{\overrightarrow{P}_{0:\zeta}}
\def\argmin{\textrm{argmin}}

We first prove \cref{prop:conv_SFBD}, which we restate below:
\convSFBD*

To facilitate our discussions, let
\begin{itemize}
	\item $\bwdM^{\phiv_{k-1}}$: the path measure induced by the backward process \cref{eq:conv_SFBD:bwd}. In general, we use $\bwdM^{\phiv}$ to denote the path measure when the drift term is parameterized $\phiv$. 
	\item $\fwdM^{(k)}$: the path measure induced by the forward process \cref{eq:fwd_diff} with $p_0 = p_0^{(k)}$, defined in \cref{prop:conv_SFBD}. The density of its marginal distribution at time $t$ is denoted by $p_t^{(k)}$
	\item $\fwdM^*$: the path measure induced by the forward process \cref{eq:fwd_diff} with $p_0 = p_\text{data}$. 
\end{itemize}
We note that, according to \cref{alg:SFBD}, the marginal distribution of $\bwdM^{\phiv_{k-1}}$ at $t = 0$ has density $p_0^{(k)}$. 

The following lemma allows us to show that the training of the diffusion model can be seen as a process of minimizing the KL divergence of two path measures. 
\begin{lemma}[\citealt{PavonA1991}, \citealt{VargasTLL2021}]
\label{appx:lem:kl_two_path}
	Given two SDEs:
	\begin{align}
		\diff \xv_t = \fv_i(\xv_t, t) \diff t + g(t) \diff \wv_t,~~~\xv_0\sim p_0^{(i)}(\xv)~~~~ t \in [0,T]
	\end{align}
	for $i = 1, 2$. Let $P^{(i)}_{0:T}$, for $i = 1, 2$, be the path measure induced by them, respectively. Then we have,
	\begin{align}
		\KL{P^{(1)}_{0:T}}{P^{(2)}_{0:T}} = \KL{p_0^{(1)}}{p_0^{(2)}} \; + \; \Eb_{P^{(1)}_{0:T}}\Bigl[\int_0^T \frac{1}{2\,g(t)^2}\,\|\fv_1(\xv_t,t)-\fv_2(\xv_t,t)\|^2\,dt\Bigr]. 
	\end{align}
	In addition, the same result applies to a pair of backward SDEs as well, where $p_0^{(i)}$ is replaced with $p_T^{(i)}$. 
\end{lemma}
\begin{proof}
	By the disintegration theorem~(e.g., see \citealt[Appx B]{VargasTLL2021}), we have 
	 \begin{align}
	 	\KL{P_1}{P_2} = \KL{p_0^{(1)}}{p_0^{(2)}} \; + \; \Eb_{P^{(1)}_{0:T}}\left[\log \frac{\diff P^{(1)}_{0:T}(\cdot | \xv_0))}{\diff P^{(2)}_{0:T}(\cdot | \xv_0)} \right],
	 \end{align}
	 where $P^{(i)}_{0:T}(\cdot |\xv_0)$ is the conditioned path measure of $P^{(i)}_{0:T}$ given the initial point $\xv_0$. Then, applying the Girsanov theorem \citep{Kailath1971, Oksendal2003} on the second term yields the desired result. 
\end{proof}
By \cref{appx:lem:kl_two_path}, we can show that the Denoiser Update step in \cref{alg:SFBD} finds $\phiv_{k}$ minimizing $\KL{\fwdM^{(k)}}{\bwdM^{\phiv}}$. To see this, note that
\begin{align*}
	\phiv_{k} &= \underset{\phiv}{\argmin} ~ \KL{\fwdM^{(k)}}{\bwdM^{\phiv}} \\
	&= \underset{\phiv}{\argmin} ~ \KL{p_\zeta^{(k)}}{p_\zeta^*} \; + \; \Eb_{\fwdM^{(k)}}\Bigl[\int_0^\zeta \frac{g(t)^2}{2}\,\| \nabla \log p_t^{(k)}(\xv_t)- \sv_{\phiv}(\xv_t,t)\|^2\,dt\Bigr],\numberthis \label{appx:eq:min_KL_update_denoiser}
\end{align*}
where $p_t^{(k)}$ is the marginal distribution induced by the forward process \eqref{eq:fwd_diff} with the boundary condition $p^{(k)}_0$ at $t = 0$. Note that, we have applied \cref{appx:lem:kl_two_path} to the backward processes inducing $\fwdM^{(k)}$ and $\bwdM^{\phiv}$. Thus, the drift term of $\fwdM^{(k)}$ is not zero but $-g(t)^2 \, \nabla \log p_t^{(k)}(\xv_t)$ according to \cref{eq:anderson_bwd}. Since the first term of \cref{appx:eq:min_KL_update_denoiser} is a constant, the minimization results in 
\begin{align}
	\nabla \log p_t^{(k)} (\xv_t) = \sv_{{\phiv}_k}(\xv_t, t)\label{eq:appx:score_minimize_KL}
\end{align}
for all $\xv_t \in \Rb^d$ and $t \in (0,\zeta]$. In addition, we note that, the denoising loss in \cref{eq:denoiser_loss} is minimized when $\nabla \log p_t^{(k)} (\xv_t) = \sv_{\phiv}(\xv_t, t)$ for all $t > 0$; as a result, $\phiv_{k}$ minimizes $\KL{\fwdM^{(k)}}{\bwdM^{\phiv}}$ as claimed. 

Now, we are ready to prove \cref{prop:conv_SFBD}. 
\begin{proof}[Proof of \cref{prop:conv_SFBD}]
Applying \cref{appx:lem:kl_two_path} to the backward process 
\begin{align*}
	\KL{\fwdM^*}{\bwdM^{\phiv_{k-1}}} &= \underbrace{\KL{p^*_\zeta}{p^*_\zeta}}_{=0} + \Eb_{\fwdM^*}\left[\int_0^\zeta \tfrac{g(t)^2}{2} \|\nabla\log p_t^*(\xv_t) - \sv_{{\phiv}_{k-1}}(\xv_t, t) \|^2 \diff t \right]\\
	&= \Eb_{\fwdM^*}\left[\int_0^\zeta \tfrac{g(t)^2}{2} \|\nabla\log p_t^*(\xv_t) - \sv_{{\phiv}_{k-1}}(\xv_t, t) \|^2 \diff t\right] \numberthis \label{eq:conv_SFBD:proof:1}
\end{align*}
Likewise,
\begin{align*}
	\KL{\fwdM^*}{\fwdM^{(k)}} =~& \KL{p^*_\zeta}{p_\zeta^{(k)}} + \Eb_{\fwdM^*}\left[\int_0^\zeta \tfrac{g(t)^2}{2} \|\nabla \log p_t^*(\xv_t) - \nabla\log p^{(k)}_t(\xv_t) \|^2 \diff t\right] \\
	=~&\KL{p^*_\zeta}{p_\zeta^{(k)}} + \Eb_{\fwdM^*}\left[\int_0^\zeta \tfrac{g(t)^2}{2} \|\nabla \log p_t^*(\xv_t) - \sv_{{\phiv}_k}(\xv_t, t) \|^2 \diff t\right] \\
	\overset{\eqref{eq:conv_SFBD:proof:1}}{=}~& \KL{p^*_\zeta}{p_\zeta^{(k)}} + \KL{\fwdM^*}{\bwdM^{\phiv_{k}}} \numberthis\label{eq:conv_SFBD:proof:2}
\end{align*} 
where the second equality is due to the discussion on deriving \cref{eq:appx:score_minimize_KL}.

\cref{appx:lem:kl_two_path} also implies that 
\begin{align}
	\KL{\fwdM^*}{\bwdM^{\phiv_{k-1}}} = \KL{p_\text{data}}{p_0^{(k)}} + \underbrace{\Eb_{\fwdM^*}\left[\int_0^\zeta \tfrac{1}{2} \|\bv^{(k-1)}(\xv_t, t) \|^2 \diff t \right]}_{:= \Bc_{k-1}}, \label{eq:conv_SFBD:proof:2_1}
\end{align}
where $\bv^{(k-1)}(\xv_t, t)$ is the drift of the forward process inducing $\bwdM^{\phiv_{k-1}}$.
In addition, 
\begin{align}
	\KL{\fwdM^*}{\fwdM^{(k)}} = \KL{p_\text{data}}{p_0^{(k)}} + \Eb_{\fwdM^*}\left[\int_0^\zeta \tfrac{1}{2} \|\zero - \zero \|^2 \diff t\right] = \KL{p_\text{data}}{p_0^{(k)}}. \label{eq:conv_SFBD:proof:3}
\end{align}
As a result, 
\begin{align*}
	\KL{p_\text{data}}{p_0^{(k)}}
	\overset{\eqref{eq:conv_SFBD:proof:3}}{=}& \KL{\fwdM^*}{\fwdM^{(k)}} 	
	\overset{\eqref{eq:conv_SFBD:proof:2}}{=} \KL{p^*_\zeta}{p_\zeta^{(k)}} + \KL{\fwdM^*}{\bwdM^{\phiv_{k}}} \\
	\geq & ~ \KL{\fwdM^*}{\bwdM^{\phiv_{k}}}
	\overset{\eqref{eq:conv_SFBD:proof:2_1}}{=} \KL{p_\text{data}}{p_0^{(k+1)}} + \Bc_k \\
	\geq & \KL{p_\text{data}}{p_0^{(k+1)}}
\end{align*}
which is \eqref{eq:conv_SFBD:monotone}. In addition, we have
\begin{align*}
	\KL{\fwdM^*}{\bwdM^{\phiv_{k-1}}} 
	&\overset{\eqref{eq:conv_SFBD:proof:2_1}}{=} \KL{p_\text{data}}{p_0^{(k)}} + \Bc_{k-1}
	\overset{\eqref{eq:conv_SFBD:proof:3}}{=} \KL{\fwdM^*}{\fwdM^{(k)}} + \Bc_{k-1} \\ 
	&\overset{\eqref{eq:conv_SFBD:proof:2}}{=} \KL{p^*_\zeta}{p_\zeta^{(k)}} + \KL{\fwdM^*}{\bwdM^{\phiv_{k}}}  + \Bc_{k-1} \\
	&= \KL{\fwdM^*}{\bwdM^{\phiv_{k}}}  + \big[\KL{p^*_\zeta}{p_\zeta^{(k)}} +  \Bc_{k-1}\big].                                                                                                                                               
\end{align*}

As a result, applying this relationship recursively, we have
\begin{align}
	\KL{\fwdM^*}{\bwdM^{\phiv_{0}}} = \sum_{k = 1}^K \KL{p^*_\zeta}{p_\zeta^{(k)}} + \sum_{k=1}^{K} \Bc_{k-1} + \KL{\fwdM^*}{\bwdM^{\phiv_{K}}}. 
\end{align}
Since $\KL{\fwdM^*}{\bwdM^{\phiv_{0}}} = M_0$, we have
\begin{align}
	\sum_{k = 1}^K \KL{p_\text{data} * h}{p^{(k)} * h} = \sum_{k = 1}^K \KL{p^*_\zeta}{p_\zeta^{(k)}} \leq M_0,
\end{align}
for all $K \geq 1$. This further implies, 
\begin{align}
	\min_{k \in \{1, 2, \ldots, K\}}\KL{p_\text{data} * h}{p^{(k)} * h} \leq \frac{M_0}{K}. 
\end{align}

Applying \cref{prop:conv_identify}, we obtain,
\begin{align}
			\min_{k \in \{1, 2, \ldots, K\}} \left|\Phi_{p_\text{data}}(\uv) - \Phi_{p_0^{(k)}}(\uv)\right|\leq \exp \big(\frac{\sigma_\zeta^2}{2} \|\uv\|^2 \big) \sqrt{\frac{2 M_0}{K}}.
\end{align}
\end{proof}

\textit{Additional Comments on the Convergence Guarantee of \cref{prop:conv_SFBD}.}
In the main text, we noted that although the bound appears to grow with $\|\mathbf{u}\|$, the behaviour of characteristic functions at large $\|\mathbf{u}\|$ is typically negligible in practice. Characteristic functions of distributions with smooth, bounded densities tend to decay rapidly -- often exponentially or at a super-polynomial rate.

To make this precise, consider the 1D case: the characteristic function $\phi(u)$ is the Fourier transform of the density. If the density is $k$-times differentiable, then it is well known (e.g., Lemma 4, p.~514, \citealt{Feller1971}) that $|\phi(u)| = o(|u|^{-k})$. This implies that for sufficiently large $|u|$, the characteristic function becomes negligible in magnitude.

Thus, assuming both $p_{\rm data}$ and $p_0^{(k)}$ are smooth with bounded support, it suffices to match their characteristic functions over a compact domain $|u| < U$ for some $U > 0$. Such local agreement in the Fourier domain implies closeness of the corresponding densities, and hence the distributions.

We complete this section by showing the connection between our framework and the original consistency loss. 
\RELCONSSFBD*
\begin{proof}
	When $t = s$, denoising noise in \cref{eq:denoiser_loss} becomes	
	\begin{align*}
	 &\underset{p_0}{\Eb} \,  \underset{p_{s|0}}{\Eb} \left[\|D_{\phiv}(\xv_s, s) - \xv_0 \|^2 \right] = \Eb_{p_s} \Eb_{p_{0|s}}\big[ \|D_{\phiv}(\xv_s, s) - \xv_0 \|^2  \big] \\
	 =\, & \Eb_{p_s} \Eb_{p_{0|s}}\big[ \|D_{\phiv}(\xv_s, s) - \Eb_{p_{0|s}}[\xv_0] + \Eb_{p_{0|s}}[\xv_0] - \xv_0 \|^2  \big] \\
	 =\, & \Eb_{p_s} \Eb_{p_{0|s}}\big[\|D_{\phiv}(\xv_s, s) - \Eb_{p_{0|s}}[\xv_0]  \|^2 \big] + \underbrace{\Eb_{p_s} \Eb_{p_{0|s}}\big[\|\Eb_{p_{0|s}}[\xv_0]  - \xv_0\|^2 \big]}_{\text{Const.}}  \\
	 &\hspace{15em} + 2 \underbrace{\Eb_{p_s} \Eb_{p_{0|s}}\big[\inner{D_{\phiv}(\xv_s, s) - \Eb_{p_{0|s}}[\xv_0]}{\Eb_{p_{0|s}}[\xv_0] - \xv_0} \big]}_{=0} \\
	 =\, & \Eb_{p_s}\big[\|D_{\phiv}(\xv_s, s) - \Eb_{p_{0|s}}[\xv_0]  \|^2 \big] + \textrm{Const.} \\
	 =\, & \Eb_{p_s} \big[\|D_{\phiv}(\xv_s, s) - \Eb_{p_{0|s}}[D_{\phiv}(\xv_0, 0)]  \|^2 \big] + \textrm{Const.},
	 \end{align*}
	 which is the consistency loss in \cref{eq:consistency_loss} when $r = 0$. 
\end{proof}

\newpage
\section{Additional Sampling Results}
\label{appx:sample_results}
In this section, we present model-generated samples used for FID computation in \cref{sec:emp}. The samples are taken from the models at their fine-tuning iteration with the lowest FID.

\subsection{CIFAR-10}
\textbf{Samples for computing FIDs in \cref{fig:ablation_cifar10:clean_ratio} - Clean Image Ratio}

\FloatBarrier

\begin{figure}[h!]
    \centering
    \includegraphics[width=0.6\linewidth]{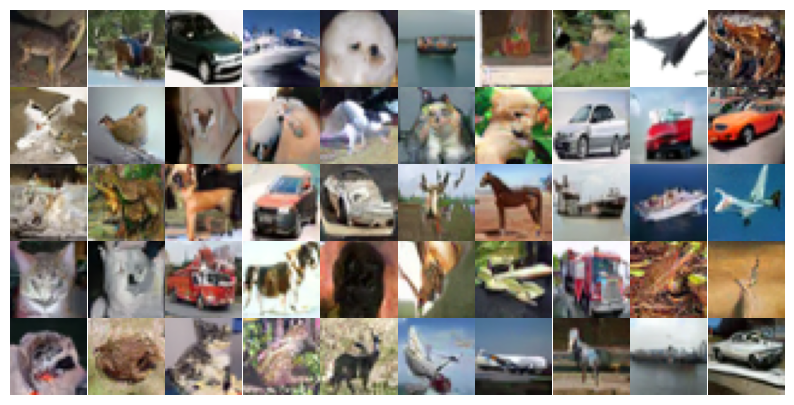}
    \caption{Clean image ratio = 0.04 -- FID: 6.31}
\end{figure}

\begin{figure}[h!]
    \centering
    \includegraphics[width=0.6\linewidth]{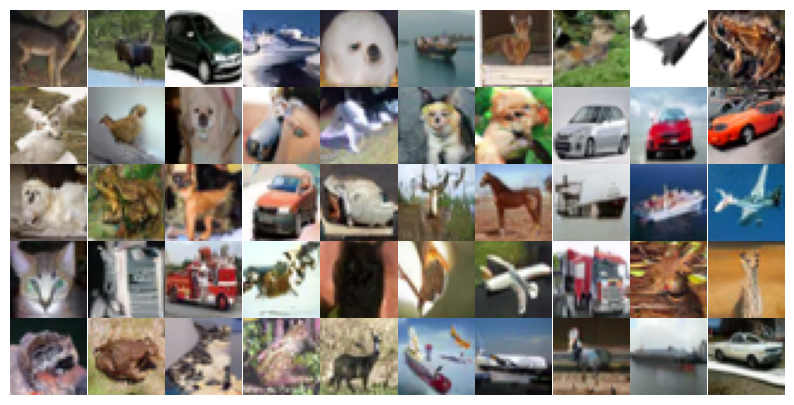}
    \caption{Clean image ratio = 0.1  -- FID: 3.58 }
\end{figure}

\begin{figure}[h!]
    \centering
    \includegraphics[width=0.6\linewidth]{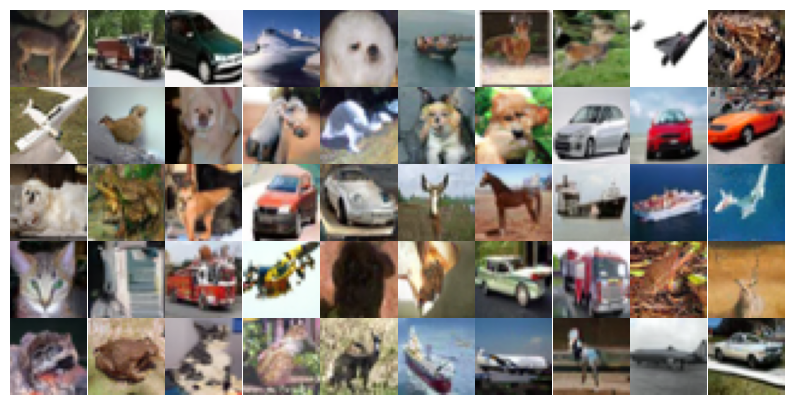}
    \caption{Clean image ratio = 0.2 -- FID: 2.98}
\end{figure}

\textbf{Samples for computing FIDs in \cref{fig:ablation_cifar10:sigma} - Noise Level}

\begin{figure}[h!]
    \centering
    \includegraphics[width=0.6\linewidth]{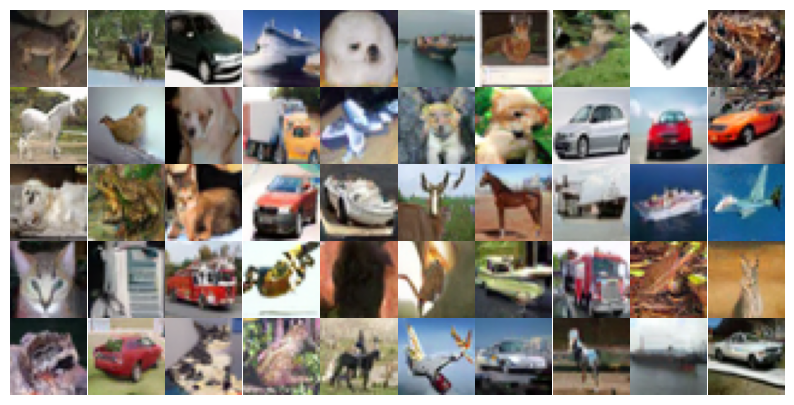}
    \caption{Noise level $\sigma_\zeta = 0.30$ -- FID: 3.97}
\end{figure}

\begin{figure}[h!]
    \centering
    \includegraphics[width=0.6\linewidth]{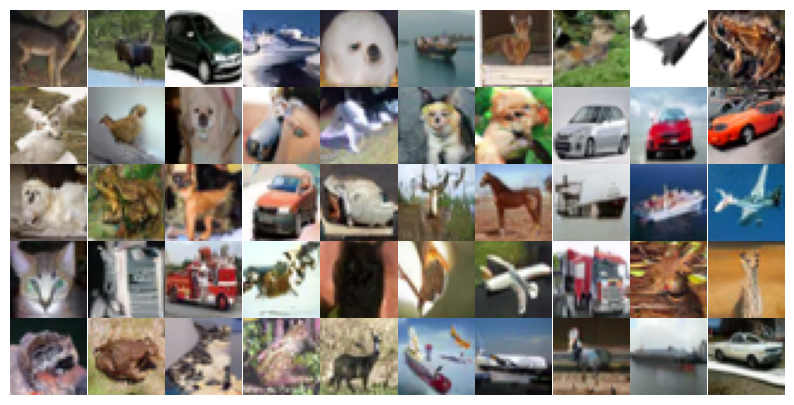}
    \caption{Noise level $\sigma_\zeta = 0.59 $ -- FID: 6.31 }
\end{figure}

\begin{figure}[h!]
    \centering
    \includegraphics[width=0.6\linewidth]{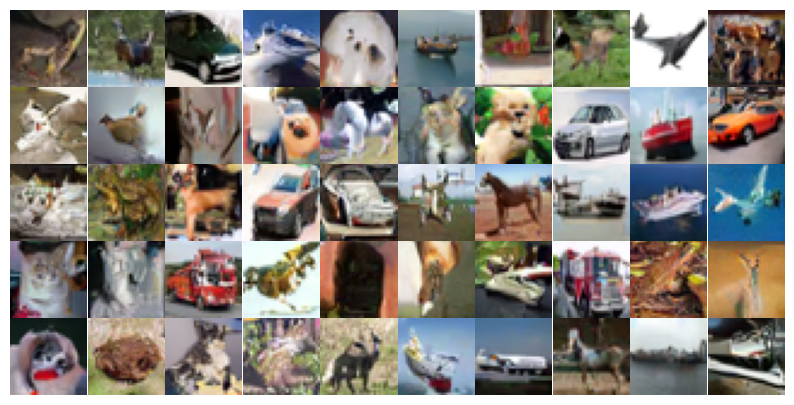}
    \caption{Noise level $\sigma_\zeta = $ 1.09 -- FID: 9.43}
\end{figure}

\begin{figure}[h!]
    \centering
    \includegraphics[width=0.6\linewidth]{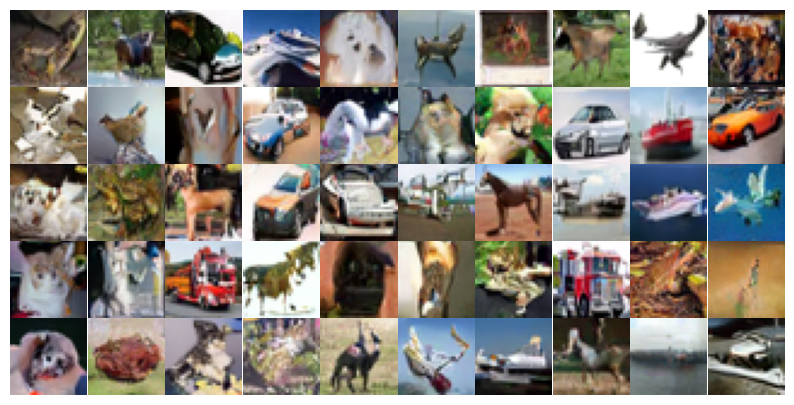}
    \caption{Noise level $\sigma_\zeta = $ 1.92 -- FID: 10.91}
\end{figure}

\FloatBarrier

\textbf{Samples for computing FIDs in \cref{fig:ablation_cifar10:pretrain} - Pretraining on Similar Datasets}

\begin{figure}[h!]
    \centering
    \includegraphics[width=0.6\linewidth]{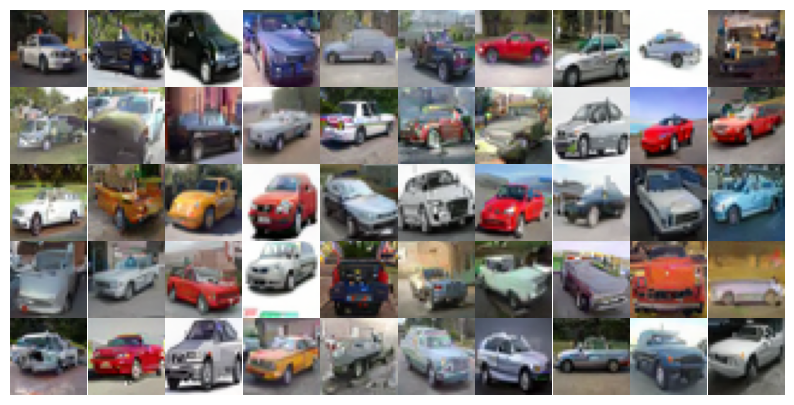}
    \caption{Class for fine-tuning: automobile -- FID: 10.39}
\end{figure}

\begin{figure}[h!]
    \centering
    \includegraphics[width=0.6\linewidth]{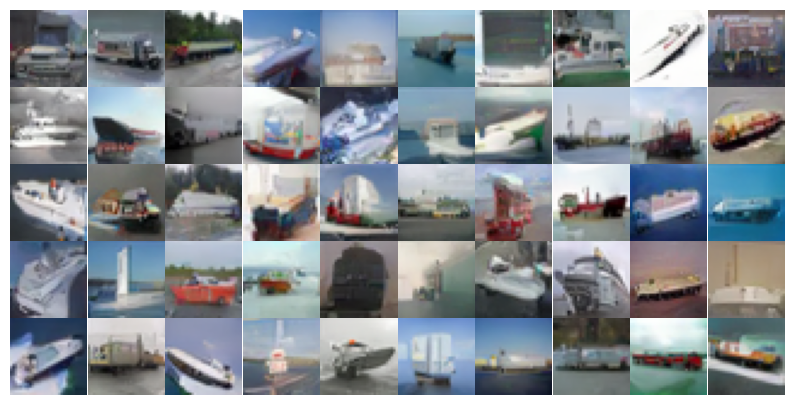}
    \caption{Class for fine-tuning: ship -- FID: 19.19}
\end{figure}

\begin{figure}[h!]
    \centering
    \includegraphics[width=0.6\linewidth]{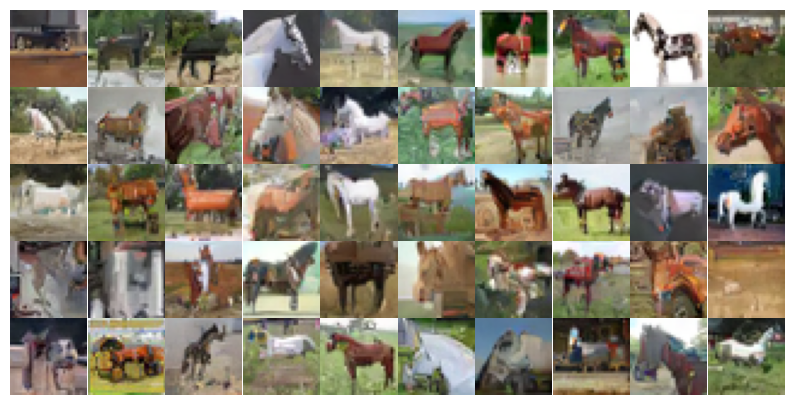}
    \caption{Class for fine-tuning: horse -- FID: 48.11}
\end{figure}
\newpage\begin{figure}[h!]
    \centering
    \includegraphics[width=0.6\linewidth]{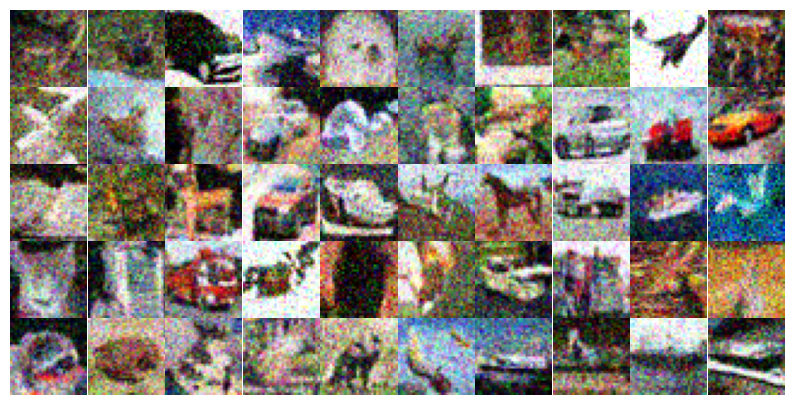}
    \caption{Class for fine-tuning: no pretrain -- FID: 155.04}
\end{figure}

\subsection{CelebA}

\begin{figure}[h!]
    \centering
    \includegraphics[width=0.8\linewidth]{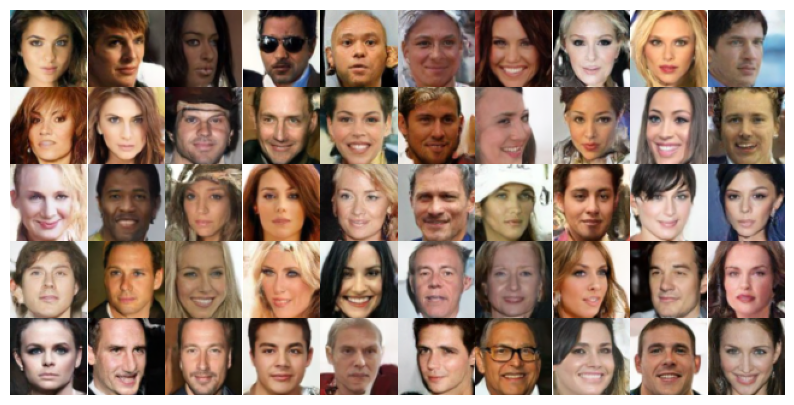}
   \caption{cfg A: $\sigma_\zeta = 1.38$; 1,500 clean images for pretraining -- FID: 5.91}
\end{figure}

\begin{figure}[h!]
    \centering
    \includegraphics[width=0.8\linewidth]{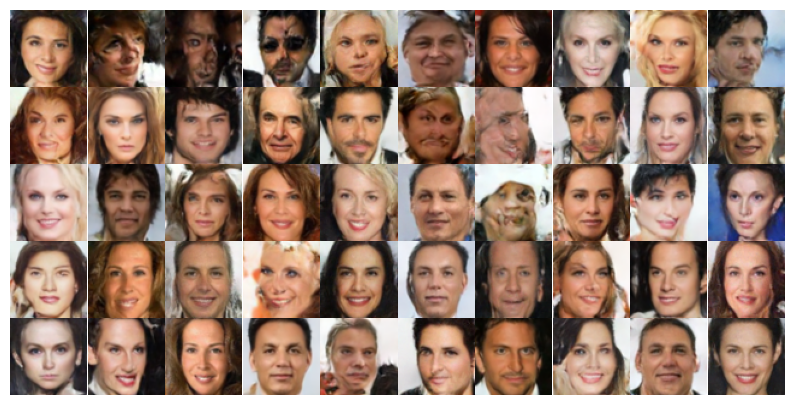}
   \caption{cfg B: $\sigma_\zeta = 1.38$; 50 clean images for pretraining -- FID: 23.63}
\end{figure}

\FloatBarrier

\begin{figure}[h!]
    \centering
    \includegraphics[width=0.8\linewidth]{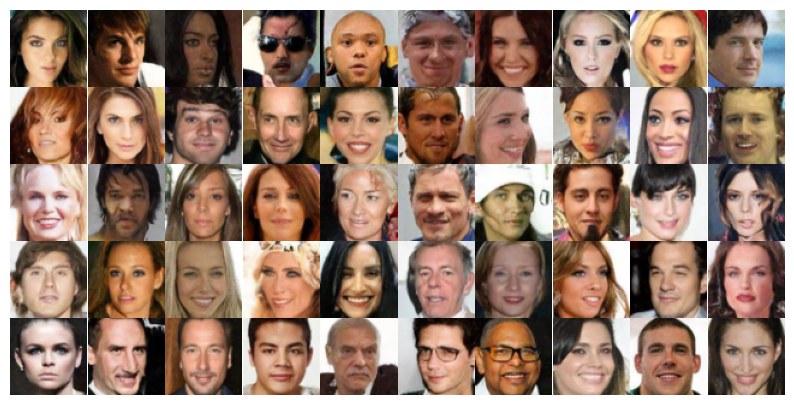}
   \caption{cfg C: $\sigma_\zeta = 0.20$; 50 clean images for pretraining -- FID: 6.48}
\end{figure}

\FloatBarrier

\newpage

\section{Experiment Configurations}
\label{appx:expConfig}
\subsection{Model Architectures}
We implemented the proposed SFBD algorithm based on the following configurations throughout our empirical studies:
\begin{table}[h]
    \centering
    \caption{Experimental Configuration for CIFAR-10 and CelebA}
    \label{tab:experiment-config}
    \renewcommand{\arraystretch}{1.2} 
    \setlength{\tabcolsep}{6pt} 
    \begin{tabular}{p{4cm} p{4.5cm} p{4.5cm}} 
        \toprule
        \textbf{Parameter} & \textbf{CIFAR-10} & \textbf{CelebA} \\
        \midrule
        \textbf{General} & & \\
        Batch Size & 512 & 256 \\
        Loss Function & \texttt{EDMLoss} \citep{KarrasAAL22} & \texttt{EDMLoss} \citep{KarrasAAL22} \\
        Sampling Method &  $2^\text{nd}$ order Heun method (EDM) \citep{KarrasAAL22} & $2^\text{nd}$ order Heun method (EDM) \citep{KarrasAAL22}  \\
        Sampling steps & 18 & 40 \\
        \midrule 
        \textbf{Network Configuration} & & \\
        Dropout & 0.13 & 0.05 \\
        Channel Multipliers & $\{2, 2, 2\}$ & $\{1, 2, 2, 2\}$ \\
        Model Channels & 128 & 128 \\
        Resample Filter & $\{1, 1\}$ & $\{1, 3, 3, 1\}$ \\
        Channel Mult Noise & 1 & 2 \\
        \midrule
        \textbf{Optimizer Configuration} & & \\
        Optimizer Class & \texttt{Adam} \citep{KingmaBa2014} & \texttt{Adam}  \citep{KingmaBa2014} \\
        Learning Rate & 0.001 & 0.0002 \\
        Epsilon & $1 \times 10^{-8}$ & $1 \times 10^{-8}$ \\
        Betas & (0.9, 0.999) & (0.9, 0.999) \\
        \bottomrule
  \end{tabular}
\end{table}

\subsection{Datasets}
All experiments on CIFAR-10 \citep{Krizhevsky2009} use only the training set, except for the one presented in \cref{fig:ablation_cifar10:pretrain}. For this specific test, we merge the training and test sets so that each class contains a total of 6,000 images. At iteration 0, the FID computation measures the distance between clean images of trucks and those from the classes on which the model is fine-tuned. For subsequent iterations, FID is calculated in the same manner as in other experiments. Specifically, the model first generates 50,000 images, and the FID is computed between the sampled images and the images from the fine-tuning classes. All experiments on CelebA \citep{LiuLWT2015} are performed on its training set.

\newpage

\section{Data Leakage and Sample Memorization}
\label{appx:data_leakage}
\newcommand{\imgdir}{figures/memorization/cifar10_pair}
As discussed in the main text, SFBD does not aim to protect the clean pretraining data from leakage. However, since these clean samples are assumed to be publicly available and free of copyright restrictions, their potential exposure poses no privacy concerns. SFBD is instead explicitly designed to safeguard \textit{sensitive} data. The key privacy-preserving mechanism is that each sensitive sample is only presented to the model in a single, corrupted form throughout training. This restriction prevents the model from memorizing or reconstructing the original content, thus reducing the risk of reproducing private or copyrighted material.

To empirically validate this claim, we adopt the evaluation protocol of \citet{DarasDD2024}, which involves analyzing similarity score distributions and identifying the most similar sensitive sample for each generated image. We report and discuss the results for both CIFAR-10 and CelebA. Overall, the findings indicate that SFBD does not regenerate sensitive examples, supporting its privacy-preserving properties. 

\subsection{CIFAR-10}

\begin{figure}[H]
	\centering
	\includegraphics[width=0.7\textwidth]{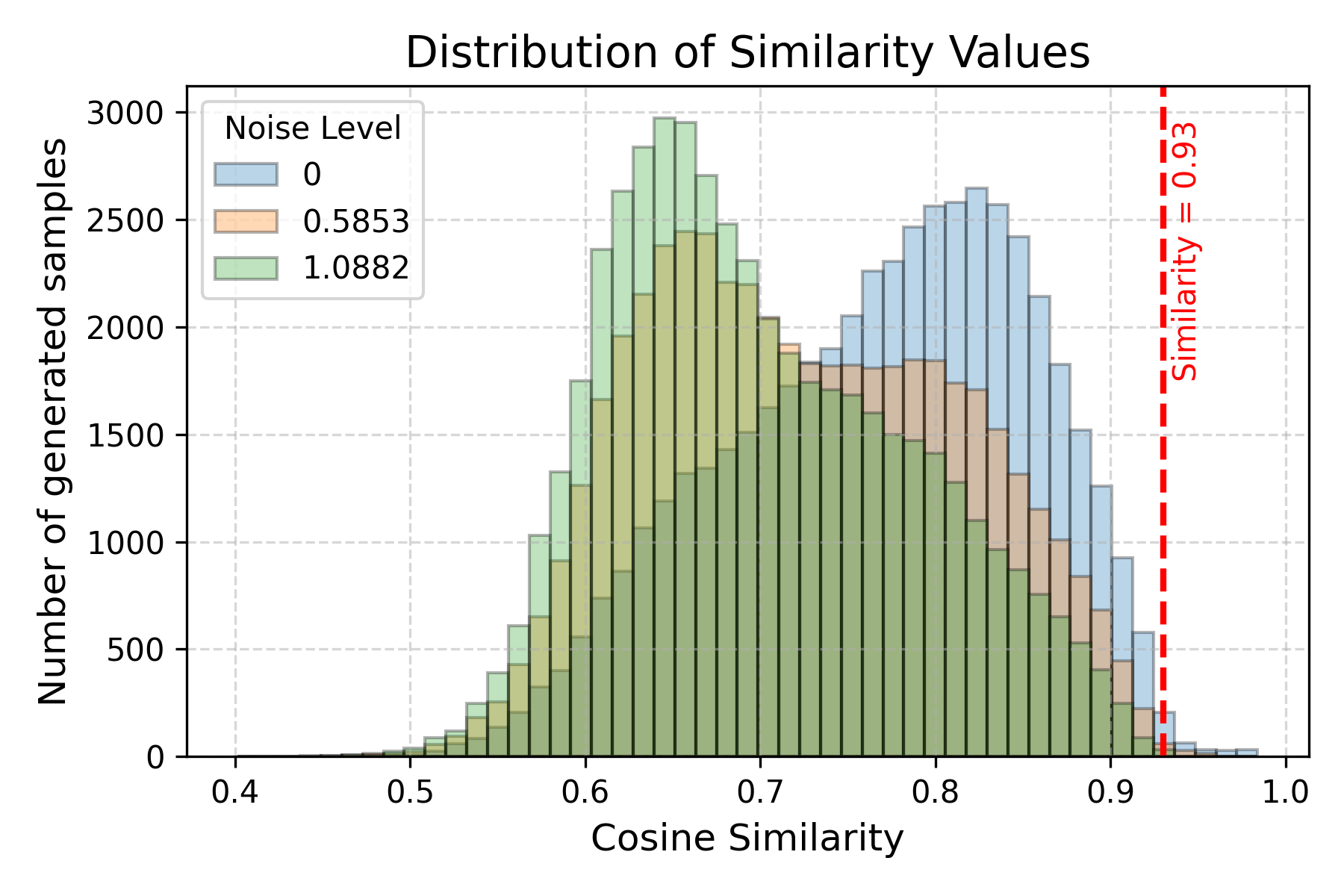}
	\caption{Distribution of maximum similarity scores for models trained with SFBD under varying noise levels. All models are pretrained using 4\% clean samples, while the remaining (sensitive) data are corrupted with noise. (The model with noise level $= 0$ is trained on the full, uncorrupted dataset.) }
	\label{fig:cifar10_similarity_value_distribution}
\end{figure}

\begin{figure}[t]
\centering

\setlength{\tabcolsep}{1pt}
\renewcommand{\arraystretch}{1.0}

\newcommand{\cifarpathwidth}{0.075\textwidth}
\begin{tabular}{ccccccc}
\includegraphics[width=\cifarpathwidth]{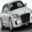} &
\includegraphics[width=\cifarpathwidth]{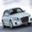} &
\includegraphics[width=\cifarpathwidth]{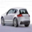} &
\includegraphics[width=\cifarpathwidth]{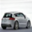} &
\includegraphics[width=\cifarpathwidth]{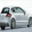} &
\includegraphics[width=\cifarpathwidth]{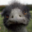} &
\includegraphics[width=\cifarpathwidth]{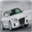} \\
\includegraphics[width=\cifarpathwidth]{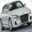} &
\includegraphics[width=\cifarpathwidth]{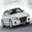} &
\includegraphics[width=\cifarpathwidth]{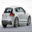} &
\includegraphics[width=\cifarpathwidth]{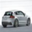} &
\includegraphics[width=\cifarpathwidth]{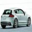} &
\includegraphics[width=\cifarpathwidth]{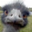} &
\includegraphics[width=\cifarpathwidth]{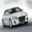} \\
0.980 & 0.969 & 0.975 & 0.973 & 0.957 & 0.949 & 0.948 \\
\end{tabular}

\vspace{1.em}

\begin{tabular}{ccccccc}
\includegraphics[width=\cifarpathwidth]{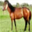} &
\includegraphics[width=\cifarpathwidth]{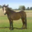} &
\includegraphics[width=\cifarpathwidth]{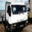} &
\includegraphics[width=\cifarpathwidth]{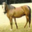} &
\includegraphics[width=\cifarpathwidth]{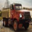} &
\includegraphics[width=\cifarpathwidth]{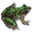} &
\includegraphics[width=\cifarpathwidth]{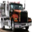} \\
\includegraphics[width=\cifarpathwidth]{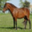} &
\includegraphics[width=\cifarpathwidth]{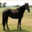} &
\includegraphics[width=\cifarpathwidth]{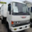} &
\includegraphics[width=\cifarpathwidth]{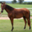} &
\includegraphics[width=\cifarpathwidth]{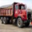} &
\includegraphics[width=\cifarpathwidth]{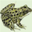} &
\includegraphics[width=\cifarpathwidth]{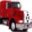} \\
0.940 & 0.930 & 0.912 & 0.918 & 0.902 & 0.901 & 0.900 \\
\end{tabular}
\caption{Generated images from the pretrained EDM on CIFAR-10 (top row) and their most similar images from the CIFAR-10 training set (bottom row). Images appear visually distinct when the similarity score falls below 0.93.}
\label{fig:cifar10_similarity_value_demo}
\end{figure}

In \cref{fig:cifar10_similarity_value_distribution}, we show the distribution of maximum similarity scores for models trained with SFBD under various noise levels. The model with noise level $= 0$ is trained on the full, uncorrupted dataset. To compute similarity, we embed both the generated images (50k samples) and the sensitive dataset images into the DINOv2 \cite{oquab2024dinov} latent space. For each generated image, we record the maximum inner product (i.e., similarity score) with its closest sensitive neighbour. As illustrated in \cref{fig:cifar10_similarity_value_demo}, images with similarity scores below 0.93 are visually distinct. Since almost all samples fall below this threshold, \cref{fig:cifar10_similarity_value_distribution} indicates that SFBD effectively avoids memorization of sensitive data while promoting sample diversity. Additionally, the figure shows that similarity scores steadily decrease as the noise level increases, supporting the privacy-preserving nature of SFBD and indicating a tradeoff between image quality and data leakage risk. 

\subsection{CelebA}

Since human faces share highly similar structures, their similarity scores are generally much higher than those observed in CIFAR-10. Consequently, instead of showing full similarity distributions, we directly present the top matching pairs between the generated images (50,000 samples) and their most similar counterparts in the sensitive dataset (used to create the noisy training set).

\begin{figure}[H]
\centering

\setlength{\tabcolsep}{2pt}
\renewcommand{\arraystretch}{0.5}

\begin{tabular}{cccccccc}
\includegraphics[width=0.1\textwidth]{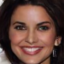} &
\includegraphics[width=0.1\textwidth]{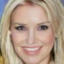} &
\includegraphics[width=0.1\textwidth]{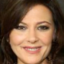} &
\includegraphics[width=0.1\textwidth]{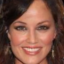} &
\includegraphics[width=0.1\textwidth]{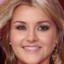} &
\includegraphics[width=0.1\textwidth]{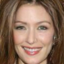} &
\includegraphics[width=0.1\textwidth]{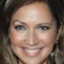} &
\includegraphics[width=0.1\textwidth]{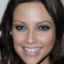} \\
\includegraphics[width=0.1\textwidth]{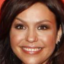} &
\includegraphics[width=0.1\textwidth]{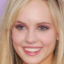} &
\includegraphics[width=0.1\textwidth]{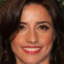} &
\includegraphics[width=0.1\textwidth]{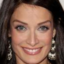} &
\includegraphics[width=0.1\textwidth]{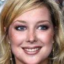} &
\includegraphics[width=0.1\textwidth]{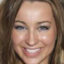} &
\includegraphics[width=0.1\textwidth]{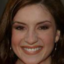} &
\includegraphics[width=0.1\textwidth]{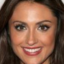} \\
0.980 & 0.978 & 0.978 & 0.978 & 0.978 & 0.977 & 0.977 & 0.977 \\
\end{tabular}

\vspace{1.5em}

\begin{tabular}{cccccccc}
\includegraphics[width=0.1\textwidth]{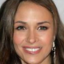} &
\includegraphics[width=0.1\textwidth]{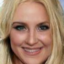} &
\includegraphics[width=0.1\textwidth]{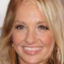} &
\includegraphics[width=0.1\textwidth]{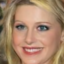} &
\includegraphics[width=0.1\textwidth]{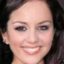} &
\includegraphics[width=0.1\textwidth]{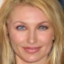} &
\includegraphics[width=0.1\textwidth]{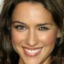} &
\includegraphics[width=0.1\textwidth]{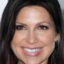} \\
\includegraphics[width=0.1\textwidth]{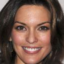} &
\includegraphics[width=0.1\textwidth]{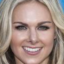} &
\includegraphics[width=0.1\textwidth]{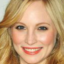} &
\includegraphics[width=0.1\textwidth]{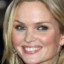} &
\includegraphics[width=0.1\textwidth]{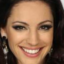} &
\includegraphics[width=0.1\textwidth]{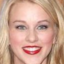} &
\includegraphics[width=0.1\textwidth]{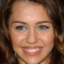} &
\includegraphics[width=0.1\textwidth]{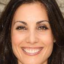} \\
0.977 & 0.977 & 0.977 & 0.976 & 0.976 & 0.976 & 0.976 & 0.976 \\
\end{tabular}

\caption{Top matching image pairs from the CelebA dataset. The model is trained in cfg A specified in the submission (1,500 clean images, $\sigma = 1.38$).  Each column shows a pair: the generated sample (top) and the most similar sensitive image (bottom) to generate noisy samples. Scores represent similarity computed via DINOv2 \citep{oquab2024dinov} and we list the top 18 pairs that give the largest similarity score.}
\label{fig:celeba_similarity_value_cfga}
\end{figure}

\begin{figure}[H]
\centering

\setlength{\tabcolsep}{2pt}
\renewcommand{\arraystretch}{0.5}

\begin{tabular}{cccccccc}
\includegraphics[width=0.1\textwidth]{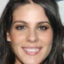} &
\includegraphics[width=0.1\textwidth]{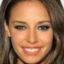} &
\includegraphics[width=0.1\textwidth]{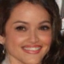} &
\includegraphics[width=0.1\textwidth]{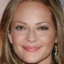} &
\includegraphics[width=0.1\textwidth]{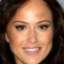} &
\includegraphics[width=0.1\textwidth]{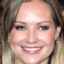} &
\includegraphics[width=0.1\textwidth]{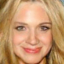} &
\includegraphics[width=0.1\textwidth]{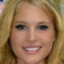} \\
\includegraphics[width=0.1\textwidth]{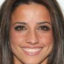} &
\includegraphics[width=0.1\textwidth]{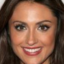} &
\includegraphics[width=0.1\textwidth]{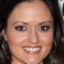} &
\includegraphics[width=0.1\textwidth]{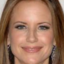} &
\includegraphics[width=0.1\textwidth]{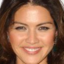} &
\includegraphics[width=0.1\textwidth]{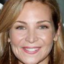} &
\includegraphics[width=0.1\textwidth]{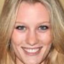} &
\includegraphics[width=0.1\textwidth]{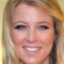} \\
0.981 & 0.980 & 0.979 & 0.979 & 0.979 & 0.979 & 0.978 & 0.978 \\
\end{tabular}

\vspace{1.5em}

\begin{tabular}{cccccccc}
\includegraphics[width=0.1\textwidth]{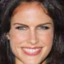} &
\includegraphics[width=0.1\textwidth]{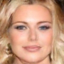} &
\includegraphics[width=0.1\textwidth]{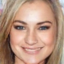} &
\includegraphics[width=0.1\textwidth]{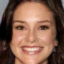} &
\includegraphics[width=0.1\textwidth]{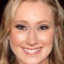} &
\includegraphics[width=0.1\textwidth]{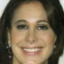} &
\includegraphics[width=0.1\textwidth]{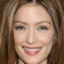} &
\includegraphics[width=0.1\textwidth]{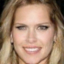} \\
\includegraphics[width=0.1\textwidth]{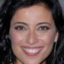} &
\includegraphics[width=0.1\textwidth]{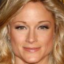} &
\includegraphics[width=0.1\textwidth]{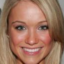} &
\includegraphics[width=0.1\textwidth]{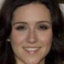} &
\includegraphics[width=0.1\textwidth]{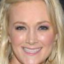} &
\includegraphics[width=0.1\textwidth]{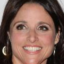} &
\includegraphics[width=0.1\textwidth]{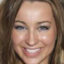} &
\includegraphics[width=0.1\textwidth]{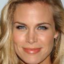} \\
0.978 & 0.978 & 0.978 & 0.978 & 0.978 & 0.978 & 0.977 & 0.977 \\
\end{tabular}
\caption{Top matching image pairs from the CelebA dataset. The model is trained in cfg C specified in the submission (50 clean images $\sigma = 0.2$).  Each column shows a pair: the generated sample (top) and the most similar sensitive image (bottom) to generate noisy samples. Scores represent similarity computed via DINOv2 \citep{oquab2024dinov} and we list the top 18 pairs that give the largest similarity score.}
\label{fig:celeba_similarity_value_cfgb}
\end{figure}

The results for cfg A and cfg C are shown in \cref{fig:celeba_similarity_value_cfga} and \cref{fig:celeba_similarity_value_cfgb}, respectively. Visual inspection confirms that the model trained with SFBD does not memorize the noisy training data.

\renewcommand{\imgdir}{figures/memorization/celeba_1.3824_top_pairs}

\renewcommand{\imgdir}{figures/memorization/celeba_0.2_top_pairs}
 
\end{document}